\documentclass[journal,11pt,draftclsnofoot,onecolumn]{IEEEtran}
\IEEEoverridecommandlockouts
\usepackage{cite}
\usepackage{amsmath,amssymb,amsfonts}
\usepackage{algorithm}
\usepackage{algorithmic}
\usepackage{graphicx}
\usepackage{textcomp}
\usepackage{xcolor}
\usepackage{enumitem}
\usepackage{url}
\def\BibTeX{{\rm B\kern-.05em{\sc i\kern-.025em b}\kern-.08em
    T\kern-.1667em\lower.7ex\hbox{E}\kern-.125emX}}

\begin{document}

\title{Contact-Anchored Proprioceptive Odometry for Legged and Wheel-Legged Robots}

% {\footnotesize \textsuperscript{*}Note: Sub-titles are not captured in Xplore and should not be used}

\author{\IEEEauthorblockN{Minxing Sun\textsuperscript{1,2,3}, Yao Mao\textsuperscript{1}}
    \thanks{This research was supported by the China Scholarship Council funding. 
        \newline
        \textsuperscript{1}Institute of Optics and Electronics, Chinese Academy of Sciences, Chengdu, China, {\tt\small{sunminxing20@mails.ucas.ac.cn, maoyao@ioe.ac.cn}}
        \newline
        \textsuperscript{2}Institute for Infocomm Research (I\textsuperscript{2}R), Agency for Science, Technology and Research, Singapore.
        \newline
        \textsuperscript{3}Shenzhen Astralldynamics Technology.
    }
}

% \author{\IEEEauthorblockN{Minxing Sun\textsuperscript{1}}
% \IEEEauthorblockA{\textit{Institute of Optics and Electronics} \\
% \textit{Chinese Academy of Sciences}\\
% Chengdu, China \\
% sunminxing20@mails.ucas.ac.cn}
% \and
% \IEEEauthorblockN{Yao Mao\textsuperscript{1}}
% \IEEEauthorblockA{\textit{Institute of Optics and Electronics} \\
% \textit{Chinese Academy of Sciences}\\
% Chengdu, China \\
% maoyao@ioe.ac.cn}
% \and

\maketitle

\begin{abstract}
Reliable odometry for legged robots without cameras or LiDAR remains challenging due to IMU drift and noisy joint velocity sensing.
This paper presents a purely proprioceptive state estimator that uses only IMU and motor measurements to estimate body pose and velocity, with a unified formulation applicable to quadruped and wheel-legged robots and extensible to other legged morphologies.
The key idea is to treat each reliable contact as a kinematic anchor: joint-torque--based foot wrench estimation selects stance contacts, and the corresponding footfall records provide intermittent world-frame constraints that suppress long-term drift.
To prevent elevation drift during extended traversal, we introduce a lightweight height clustering and time-decay correction that snaps newly recorded footfall heights to previously observed support planes.
For wheel-legged platforms, the recorded contact is further propagated by effective wheel rolling displacement with shank-motion compensation and a slope-aware rolling direction.
To improve foot velocity observations under encoder quantization, we retain an inverse-kinematics cubature Kalman filter as an optional velocity-enhancement module that filters foot-end velocities from joint angles and velocities.
The implementation further mitigates yaw drift through multi-contact geometric consistency, which is injected as a soft heading prior rather than as a hard reset of the attitude state.

We evaluate the method on four quadruped platforms (three Astrall robots and a Unitree Go2 EDU) using closed-loop trajectories.
On Astrall point-foot robot~A, a $\sim$200\,m horizontal loop and a $\sim$15\,m vertical loop return with 0.1638\,m and 0.219\,m error, respectively; on wheel-legged robot~B, the corresponding errors are 0.2264\,m and 0.199\,m.
On wheel-legged robot~C, a $\sim$700\,m horizontal loop yields 7.68\,m error and a $\sim$20\,m vertical loop yields 0.540\,m error.
Unitree Go2 EDU closes a $\sim$120\,m horizontal loop with 2.2138\,m error and a $\sim$8\,m vertical loop with less than 0.1\,m vertical error.
We publicly release the complete implementation (\url{https://github.com/ShineMinxing/CAPO-LeggedRobotOdometry}) and the Go2 EDU test data through the GitHub Releases tag \texttt{DataForTest}, including ROS bags, CSV files, and videos, to enable reproducible evaluation.
\end{abstract}

\begin{IEEEkeywords}
quadruped robots, proprioceptive odometry, sensor fusion, cubature Kalman estimator
\end{IEEEkeywords}

\section{Introduction}
Accurate estimation of a legged robot's pose and velocity is a prerequisite for motion planning and stable locomotion control \cite{MV2020, PQ2021}. 
In many deployed systems, exteroceptive sensors such as depth cameras and LiDAR are coupled with SLAM to provide global state feedback \cite{FM2019, PG2022, DC2022}. 
However, these sensors can become unreliable in the presence of challenging illumination, airborne particulates, or vegetation clutter, and their computational and installation requirements may be undesirable for lightweight or cost-sensitive platforms. 
This motivates proprioceptive state estimation methods that remain effective when no vision or LiDAR feedback is available.

Most legged robots provide rich proprioceptive measurements, including an inertial measurement unit (IMU) and motor encoder readings; many platforms additionally provide estimated joint torques or contact-related signals. 
Early approaches relied heavily on IMU acceleration integration (with frame transformations and additional dynamic cues) \cite{SB2013, ED2016, MY2023}. 
In practice, double integration is extremely sensitive to bias, timing mismatch, and discrete sampling noise, leading to rapid drift. 
To mitigate this, modern estimators incorporate leg kinematics to introduce velocity constraints from stance feet: when a foot is stationary relative to the ground, the body velocity can be inferred from the relative foot--body motion. 
A key difficulty is determining which legs provide valid constraints, especially under imperfect contact conditions. 
Probabilistic contact estimation and impact detection have been used to gate these measurements and detect slippage \cite{PC2017}. 
Building on contact-aware kinematic constraints, invariant EKF formulations have been explored for legged state estimation \cite{CH2020}, and learning-based measurement models have further improved contact classification under difficult terrains \cite{LY2024}. 
Hybrid systems that incorporate cameras can provide stronger observability \cite{CY2023}, but they reintroduce the fragility of exteroception that motivates proprioceptive odometry in the first place. 
Moreover, even when velocity estimation is improved, position drift can still accumulate over time, and foot velocity computed directly from noisy joint velocities may exhibit spikes caused by encoder quantization and numerical differentiation.

This work targets robust \emph{purely proprioceptive} odometry for biped, quadruped, and wheel-legged robots using only IMU and motor measurements.
The key idea is to anchor the body state to discrete footfall events: we record the world-frame footfall position at touchdown and treat it as an intermittent constraint during the stance phase. 
In our implementation, stance selection is enabled by joint-torque--based foot wrench estimation, while wheel-legged locomotion is handled by compensating the effective wheel rotation during ground contact. 
To prevent long-term elevation drift, we introduce a lightweight height clustering and time-decay correction that snaps newly recorded footfall heights to previously observed support planes within a configurable resolution. 
To address spiky velocity observations from joint encoders, we further employ a cubature Kalman filter (CKF) under an inverse-kinematics observation model to estimate and smooth foot-end velocities directly from joint angles and angular velocities. 
Compared with linear Kalman filtering and EKF-style approximations \cite{NK1960, FS2001, SZ2010, RS2020, IB1993, RZ2017} and with alternatives such as invariant EKF and UKF \cite{IB2017, CH2020, LY2024, UW2000, UJ2004, IS2024}, CKF provides a numerically stable, Jacobian-free nonlinear filtering rule and avoids UKF-style scaling parameters, which is convenient for inverse-kinematics measurements \cite{CA2009, SS2021, AL2021}. 
Finally, we correct long-horizon yaw drift by enforcing multi-contact geometric consistency between body-frame kinematics and world-frame contact records, which also provides a fallback yaw reference when IMU yaw constraints are degraded or disabled.

\textbf{Contributions.} This paper makes the following contributions:
\begin{itemize}[leftmargin=*, itemsep=0.25em, topsep=0.25em]
    \item \textbf{Unified kinematics-based proprioceptive odometry across morphologies.}
    We formulate a contact-anchored odometry framework that relies only on IMU and motor measurements (joint angles, velocities, and estimated torques).
    The framework is morphology-agnostic: at each time step, it forms constraints only from the currently contacting end-effectors (feet or wheels), so the same estimator applies to bipeds, quadrupeds, and wheel-legged robots with different numbers of contacts over time.
    In our implementation, stance is selected via joint-torque--based foot wrench estimation, while point-foot and wheel-legged platforms share a chain-based kinematic observation model; wheel-legged platforms are additionally supported through effective rolling-contact propagation during stance.

    \item \textbf{Footfall recording with height correction for accurate elevation.}
    We record world-frame contact (footfall) positions at touchdown and use them as intermittent constraints during stance to suppress long-horizon drift.
    To prevent elevation drift over extended traversal, we propose a lightweight support-plane height clustering with time-decayed confidence that snaps newly recorded footfall heights to previously observed planes at a configurable resolution.

    \item \textbf{Inverse-kinematics nonlinear filtering for encoder-induced velocity spikes.}
    To mitigate spiky foot velocity observations caused by encoder quantization and discrete differentiation, we retain an inverse-kinematics CKF (IKVel-CKF) that estimates foot-end velocities directly from joint angles and angular velocities under a nonlinear observation model and can be enabled as CAPO-CKE before the contact-anchored fusion update.

    \item \textbf{Yaw drift suppression with a kinematics-only fallback.}
    We reduce long-horizon yaw drift by enforcing multi-contact geometric consistency between body-frame kinematics and world-frame contact records.
    In particular, during prolonged standing with stable multi-contact, this constraint effectively arrests IMU yaw drift by continuously re-anchoring heading to the contact geometry.
    The released estimator applies this correction as a soft bias on the yaw observation entering the attitude estimator; when IMU yaw constraints are disabled, the same geometry still provides a kinematics-derived heading reference, although residual drift remains sensitive to unmodeled attitude coupling and contact compliance.

    \item \textbf{Real-robot evaluation on multiple platforms and public artifacts.}
    We evaluate the estimator on four platforms (three Astrall robots and a Unitree Go2 EDU) using long closed-loop trajectories with both horizontal and vertical motion.
    We publicly release the complete implementation and representative Go2 EDU test data through the \texttt{DataForTest} GitHub release, including ROS bags, CSV files, and videos, to enable reproducible evaluation.
\end{itemize}

The remainder of this paper is organized as follows:
Section~II presents the contact-anchored proprioceptive odometry formulation, including stance selection, footfall recording, and the support-plane height correction strategy.
Section~III discusses end-effector radius modeling and wheel contact propagation, and analyzes the approximation error introduced when modeling a rounded point-foot as an effective shank extension.
Section~IV introduces the IKVel-CKF used to suppress encoder-induced velocity spikes and to provide smoother hip--foot velocity feedback for fusion.
Section~V presents a kinematics-based yaw estimation method based on multi-contact geometric consistency, and discusses its ability to arrest IMU yaw drift during prolonged standing as well as its graceful degradation when IMU yaw constraints are unreliable.
Section~VI reports evaluation results in both physics simulation and real-robot experiments, covering flat-ground and stair scenarios as well as long closed-loop trajectories on multiple platforms.
Section~VII concludes the paper and outlines future work.

\section{Contact-Anchored Proprioceptive Odometry}
\label{sec:contact_anchored}

This section presents the contact-anchored proprioceptive odometry implemented in our system, including stance selection, footfall recording, and support-plane height correction. Wheel-specific contact propagation is addressed separately in Section~III.

\subsection{Inputs, Frames, and Notation}
\label{sec:sec2_notation}
We consider a world frame $\mathcal{F}_W$ and a body frame $\mathcal{F}_B$ attached to the robot trunk.
At time $t_k$, the trunk pose is represented by position $\mathbf{p}_{WB,k}\in\mathbb{R}^3$ and rotation $\mathbf{R}_{WB,k}\in\mathrm{SO}(3)$.
For each leg $i$, motor measurements provide joint angles $\mathbf{q}_{i,k}$, joint angular velocities $\dot{\mathbf{q}}_{i,k}$, and estimated joint torques $\boldsymbol{\tau}_{i,k}$.
Robot geometry provides the hip mounting position $\mathbf{p}^{B}_{\mathrm{hip},i}$ and the link or transform-chain parameters required by the forward kinematics.

We use right-handed frames with $x$ forward, $y$ left, and $z$ upward.
The estimated end-effector force $\mathbf{f}$ is the wrench-equivalent force consistent with the virtual-work mapping
$\boldsymbol{\tau}=\mathbf{J}^\top\mathbf{f}$ (Appendix~\ref{app:leg_kin_wrench}).
Under our sign convention, a supporting contact produces a negative vertical component.

We denote the set of contacting end-effectors (feet or wheels) at time $t_k$ as $\mathcal{C}_k$.
For each leg, the chain kinematics yield the end-effector position and the joint-rate-induced end-effector velocity in the world frame,
and wrench estimation yields a contact-normal force used for stance gating.
The explicit kinematic and wrench formulas, including the transform-chain form used in the released implementation and the 3-DoF closed-form specialization used for analysis, are given in Appendix~\ref{app:leg_kin_wrench}.

\subsection{Stance Selection via Torque-Based Wrench Estimation}
\label{sec:sec2_stance}
To enable contact-anchored constraints, we must select legs that are in reliable ground contact.
In our implementation, the end-effector force is estimated from joint torques and the geometric Jacobian (Appendix~\ref{app:leg_kin_wrench}).
Let $\mathbf{f}^{W}_{i,k}$ denote the end-effector force expressed in the world-aligned frame used by the stance gate.
In the implementation, this force is obtained directly from the world-frame transform-chain Jacobian and the measured joint torques, rather than by first forming a separate body-frame wrench and then explicitly rotating it.
We declare leg $i$ as contacting when the vertical component satisfies
\begin{equation}
\label{eq:contact_gate2}
i\in\mathcal{C}_k \quad \Leftrightarrow \quad f^{W}_{i,k,z} < f_{\mathrm{th}},
\end{equation}
where $f_{\mathrm{th}}$ is a configurable threshold under our sign convention.
For weighting auxiliary geometric cues and for diagnostics, we also compute a soft contact confidence
\begin{equation}
\label{eq:contact_probability}
\rho_{i,k}=
\begin{cases}
0, & f^{W}_{i,k,z}\ge 0.3f_{\mathrm{th}},\\
1, & f^{W}_{i,k,z}\le 1.3f_{\mathrm{th}},\\
\dfrac{f^{W}_{i,k,z}-0.3f_{\mathrm{th}}}{f_{\mathrm{th}}}, & \mathrm{otherwise},
\end{cases}
\end{equation}
which is consistent with the negative-force contact convention and provides a smooth transition between swing and stance.

A touchdown event is detected by
\begin{equation}
\label{eq:touchdown2}
\mathrm{Touchdown}(i,k)\ \Leftrightarrow\ \big(i\in\mathcal{C}_k\big)\wedge\big(i\notin\mathcal{C}_{k-1}\big),
\end{equation}
which triggers footfall recording and height correction (Section~\ref{sec:sec2_height}).

\subsection{Footfall Recording and Contact-Anchored Body Observations}
\label{sec:sec2_anchor}

\begin{figure}[htbp]
    \centering
    \begin{minipage}[b]{0.48\linewidth}
        \centering
        \includegraphics[width=1\linewidth]{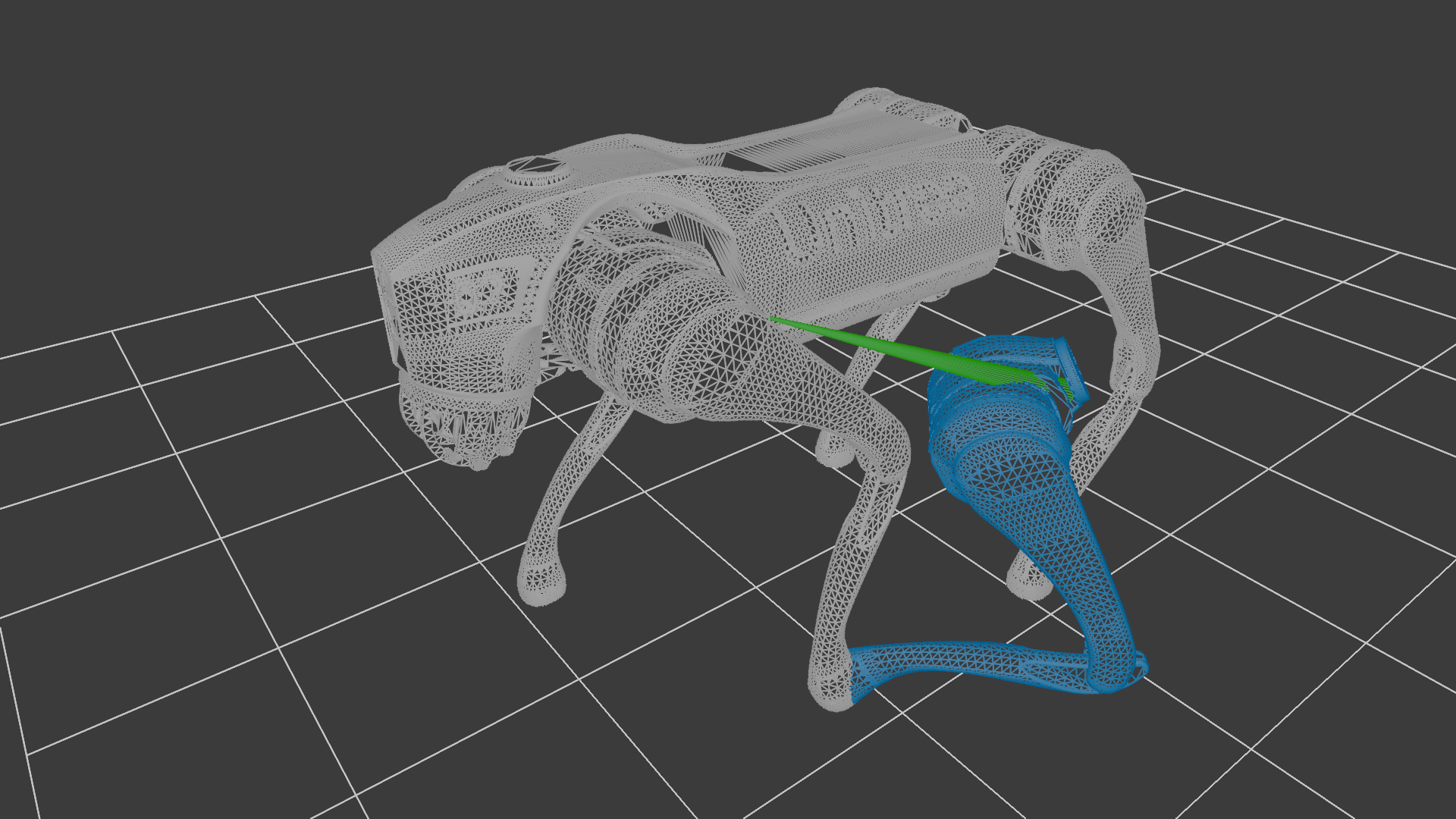}
        \caption{Footfall records provide continuous body position feedback.}
        \label{P2_3_1}
    \end{minipage}
    \hfill
    \begin{minipage}[b]{0.48\linewidth}
        \centering
        \includegraphics[width=1\linewidth]{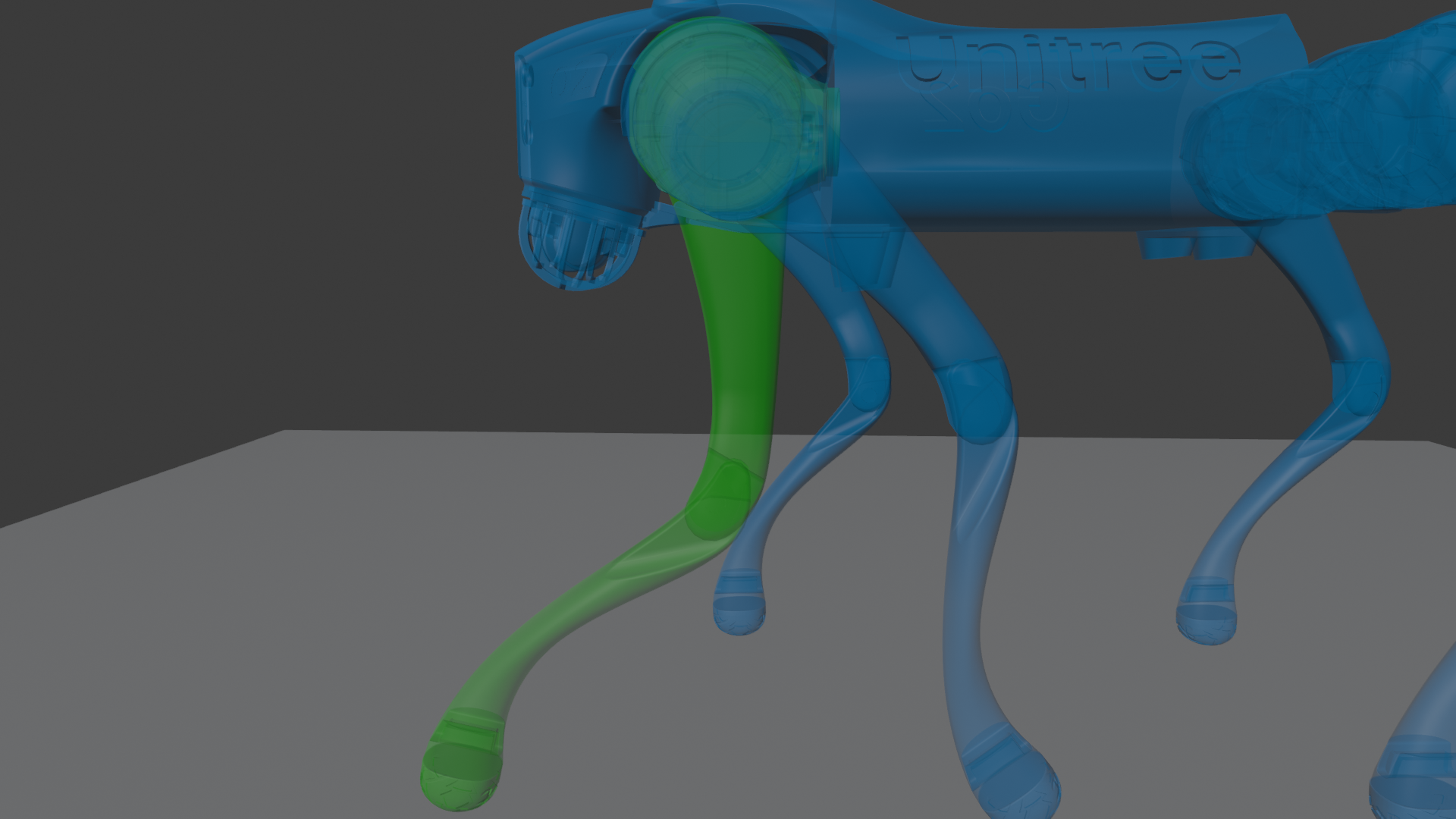}
        \caption{Encoder-derived joint rates provide continuous body velocity feedback.}
        \label{P2_3_2}
    \end{minipage}
\end{figure}

Given leg $i$, forward kinematics provides the hip-to-end-effector vector in the body frame,
$\mathbf{r}^{B}_{i,k}$, and its velocity $\dot{\mathbf{r}}^{B}_{i,k}$ (Appendix~\ref{app:leg_kin_wrench}).
The body-frame end-effector position is then
\begin{equation}
\label{eq:foot_pos_body}
\mathbf{p}^{B}_{\mathrm{ee},i,k}=\mathbf{p}^{B}_{\mathrm{hip},i}+\mathbf{r}^{B}_{i,k},
\end{equation}
and its world-frame counterpart is
\begin{equation}
\label{eq:foot_pos_world}
\mathbf{p}^{W}_{\mathrm{ee},i,k}=\mathbf{p}_{WB,k}+\mathbf{R}_{WB,k}\mathbf{p}^{B}_{\mathrm{ee},i,k}.
\end{equation}

\paragraph{Footfall record.}
When $\mathrm{Touchdown}(i,k)$ is true, we initialize or reset the footfall record as
\begin{equation}
\label{eq:footfall_rec2}
\mathbf{c}^{W}_{i}\ \leftarrow\ \mathbf{p}^{W}_{\mathrm{ee},i,k}.
\end{equation}
For a previously uninitialized stance record, the implementation bootstraps the horizontal coordinates from the current body estimate and initializes the vertical coordinate by the ground-height prior; subsequent touchdown events and the height-memory mechanism then refine the full three-dimensional anchor.

Fig.~\ref{P2_3_1} illustrates the anchoring intuition: at touchdown we store the world-frame contact point $\mathbf{c}^W_i$; during stance the contact is treated as stationary, so the trunk position is obtained by subtracting the current body-frame kinematics (rotated into $\mathcal{F}_W$) from the stored footfall record.

\paragraph{Contact-anchored position observation.}
During stance ($i\in\mathcal{C}_k$), the footfall point is assumed stationary in $\mathcal{F}_W$.
Thus the trunk position implied by leg $i$ is
\begin{equation}
\label{eq:pos_obs_leg2}
\tilde{\mathbf{p}}_{WB,k}(i)=\mathbf{c}^{W}_{i}-\mathbf{R}_{WB,k}\mathbf{p}^{B}_{\mathrm{ee},i,k}.
\end{equation}

\paragraph{Contact-anchored velocity observation.}

The released estimator forms the stance velocity observation directly from the world-frame chain Jacobian.
Let $\mathbf{J}^{W}_{i,k}$ map the active joint rates of leg $i$ to the end-effector velocity induced by relative leg motion, i.e.,
\begin{equation}
\label{eq:world_jacobian_short}
\dot{\mathbf{r}}^{W}_{i,k}=\mathbf{J}^{W}_{i,k}\dot{\mathbf{q}}_{i,k}.
\end{equation}
If the stance contact is stationary, the body velocity implied by this leg is the opposite of the relative foot motion; for wheel contacts, the rolling velocity is added along the estimated rolling direction:
\begin{equation}
\label{eq:vel_obs_leg}
\tilde{\mathbf{v}}_{WB,k}(i)
=
-\mathbf{J}^{W}_{i,k}\dot{\mathbf{q}}_{i,k}
+
\mathbf{v}^{\mathrm{wheel}}_{i,k}.
\end{equation}
For point-foot robots, $\mathbf{v}^{\mathrm{wheel}}_{i,k}=\mathbf{0}$.
This formulation is equivalent to using the contact as an instantaneous velocity anchor, but it avoids introducing an additional IMU angular-velocity cross term into the leg pseudo-observation; attitude enters through the world-frame kinematic chain.
Since encoder quantization and discrete differentiation can introduce spikes, Section~IV retains IKVel-CKF as an optional replacement or smoother for the default Jacobian velocity term.
\begin{equation}
\label{eq:fuse_obs2}
\tilde{\mathbf{p}}_{WB,k}=\frac{1}{|\mathcal{C}_k|}\sum_{i\in\mathcal{C}_k}\tilde{\mathbf{p}}_{WB,k}(i),\qquad
\tilde{\mathbf{v}}_{WB,k}=\frac{1}{|\mathcal{C}_k|}\sum_{i\in\mathcal{C}_k}\tilde{\mathbf{v}}_{WB,k}(i).
\end{equation}
The fused observation $(\tilde{\mathbf{p}}_{WB,k},\tilde{\mathbf{v}}_{WB,k})$ updates the translational estimator when a reliable support set is available.
If no reliable contact can be identified, the contact-anchored update is suppressed or conservatively regularized, and the estimator relies on the remaining proprioceptive prediction and inertial observations.

\subsection{Support-Plane Height Correction}
\label{sec:sec2_height}

\begin{figure}[htbp]
    \centering
    \includegraphics[width=0.9\textwidth]{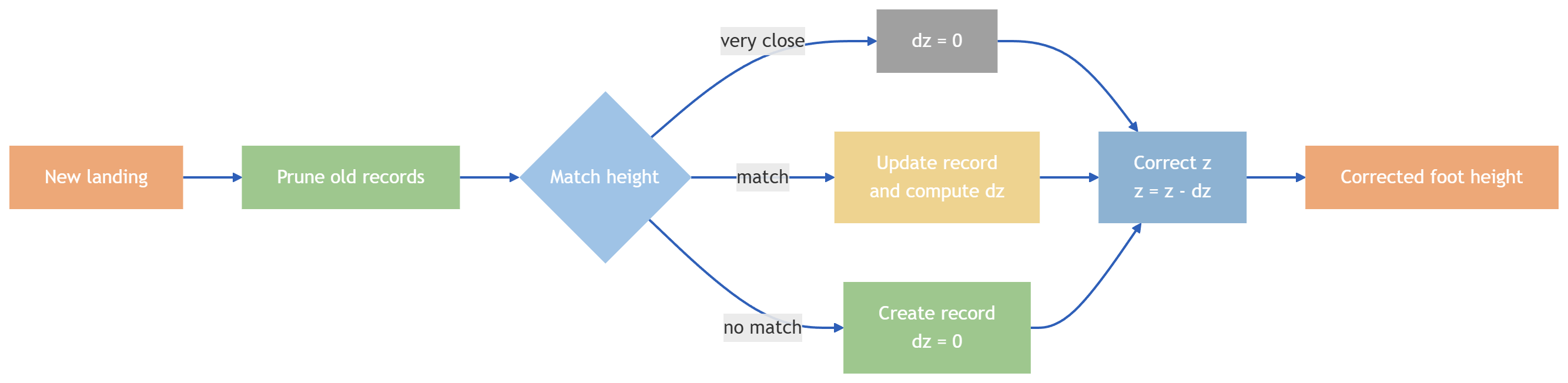}
    \caption{Footstep tracking correction strategy flowchart.}
    \label{P2_4_1}
\end{figure}
\begin{figure}[htbp]
    \centering
    \includegraphics[width=0.6\textwidth]{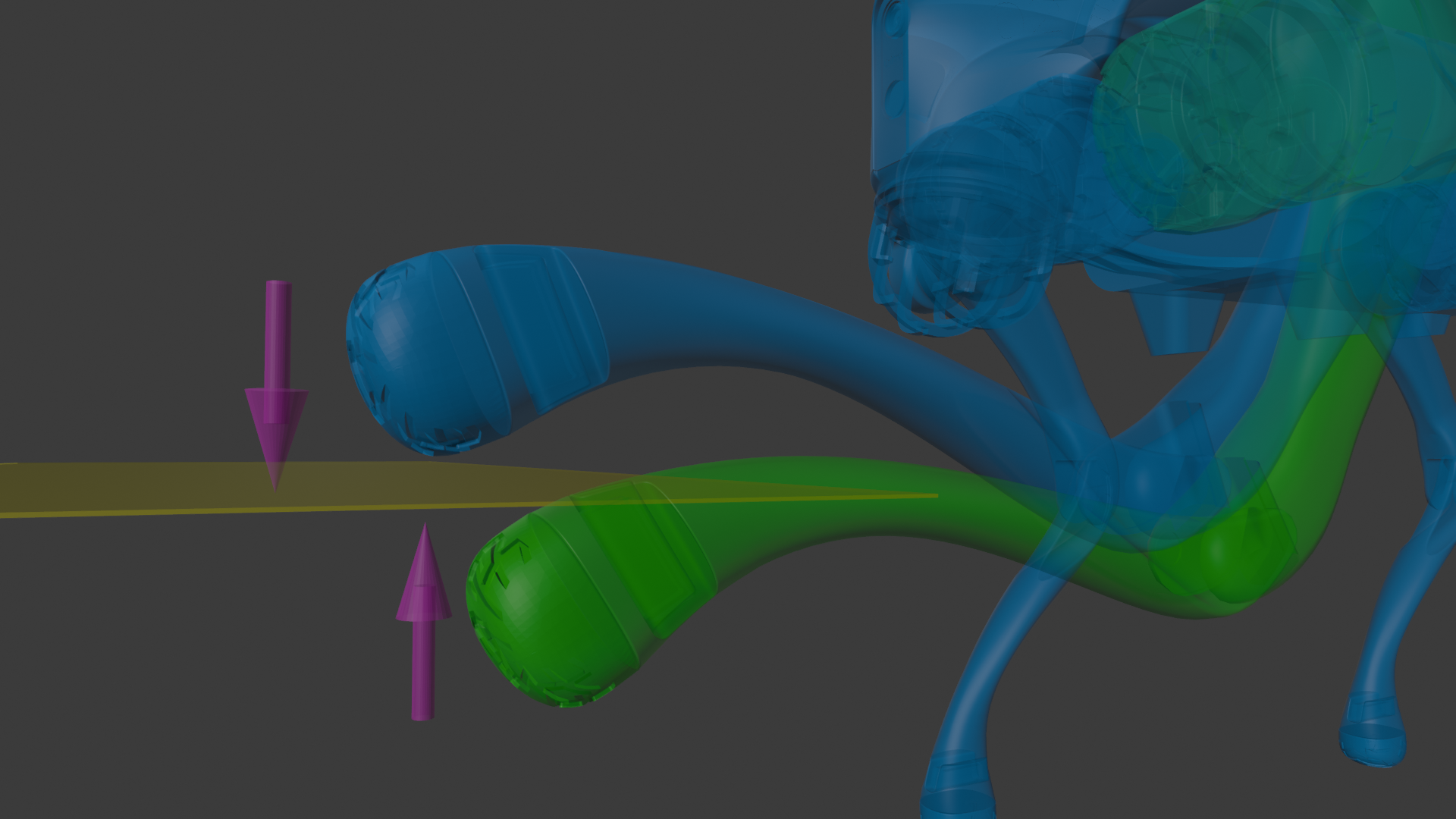}
    \caption{Correction of footstep position.}
    \label{P2_4_2}
\end{figure}

Directly recorded footfall heights are sensitive to small biases; without correction, vertical drift accumulates over long traversal.
We maintain a discrete set of support-plane height records
\begin{equation}
\label{eq:height_store2}
\mathcal{H}=\{(h_n,w_n,t_n)\}_{n=1}^{N_h},
\end{equation}
where $h_n$ is the plane height, $w_n$ a confidence weight, and $t_n$ the last update time.

Fig.~\ref{P2_4_1} summarizes the touchdown-triggered correction logic.
At each touchdown, we (i) prune stale support-plane records, (ii) associate the raw touchdown height to an existing plane if possible, and (iii) either snap-and-refresh the matched plane or create a new one.
Fig.~\ref{P2_4_2} illustrates a typical stair/step scenario where repeated small touchdown biases would otherwise accumulate into elevation drift.

Let the raw touchdown height be $z=\mathbf{c}^{W}_{i}(3)$ at time $t_k$.
We maintain a set of plane records $\mathcal{H}=\{(h_n,w_n,t_n)\}_{n=1}^{N_h}$ and first discard records with $t_k-t_n>T_{\mathrm{fade}}$.
A plane is considered a match if $|z-h_n|\le \Delta_h$; under slope-aware wheel propagation, a relaxed step-height tolerance can also be used so that gradual support-plane changes are not mistaken for unrelated terrain layers.
If a match exists, we compute
\begin{equation}
\label{eq:height_match}
\Delta z =
\begin{cases}
0, & |z-h_n|\le \Delta_h/10,\\
z-h_n, & \text{otherwise},
\end{cases}
\qquad
z \leftarrow z-\Delta z ,
\end{equation}
and update its confidence with time decay,
\begin{equation}
\label{eq:height_weight}
w_n \leftarrow w_n\exp\!\left(-\frac{t_k-t_n}{\kappa T_{\mathrm{fade}}}\right)+1,\qquad
t_n \leftarrow t_k .
\end{equation}
If no match exists, we create a new record $(h_n,w_n,t_n)\leftarrow (z,1,t_k)$.
Finally, the corrected value is written back to the footfall record, $\mathbf{c}^{W}_{i}(3)\leftarrow z$.
This correction is applied only at touchdown and stabilizes long-horizon elevation estimation under repeated height changes (Fig.~\ref{P2_4_1}).

\section{End-Effector Radius Modeling and Wheel Contact Compensation}
\label{sec:radius_wheel}

Contact-anchored proprioceptive odometry is only as accurate as the underlying end-effector contact model.
For rounded point feet, the stance contact can roll even without slip; for wheel-legged robots, the contact point propagates through wheel rotation.
This section first quantifies the systematic bias introduced when a rounded point foot is approximated as a rigid shank extension, which motivates the wheel-contact treatment developed in Section~III-B.

\subsection{Rolling-Contact Bias of Rounded Point Feet}
\label{sec:foot_radius_bias}

\begin{figure}[htbp]
    \centering
    \includegraphics[width=0.6\linewidth]{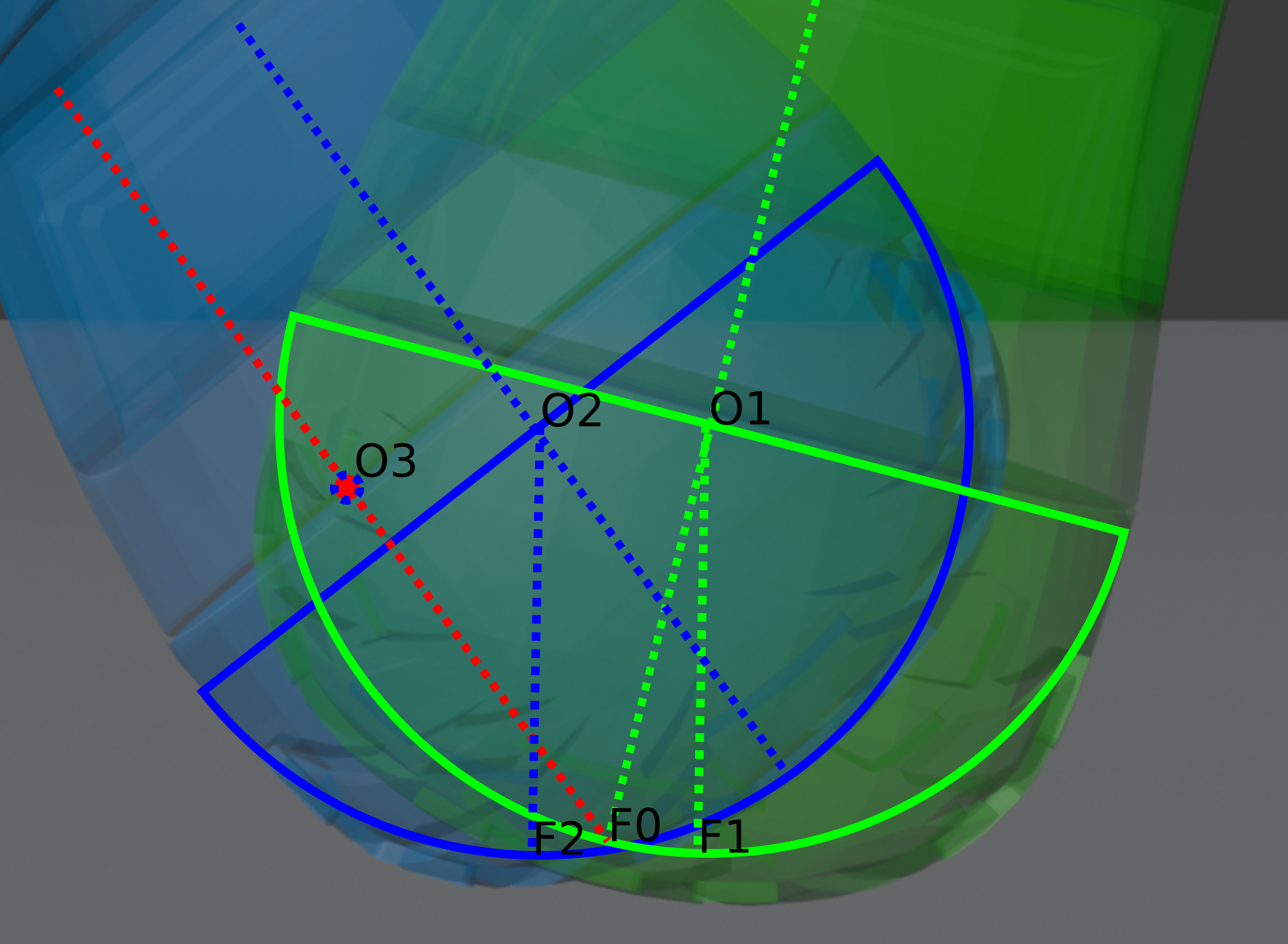}
    \caption{Sagittal-plane model of a rounded point foot. $O_1$ is the shank--foot connection. Touchdown and lift-off shank pitch angles are $a_1$ and $a_2$. Ignoring rolling yields the fictitious displacement $O_1\!\rightarrow\!O_3$, while hemispherical rolling yields the physical displacement $O_1\!\rightarrow\!O_2$.}
    \label{fig:foot_geom}
\end{figure}

We analyze a 2D sagittal-plane ($x$--$z$) model, where $x$ is forward and $z$ is upward.
The foot is modeled as a hemisphere of radius $R$.
Let $O_1$ be the shank--foot connection point at touchdown, and let $a_1$ and $a_2$ denote the shank pitch angles (measured from the $+x$ axis) at touchdown and lift-off, respectively (Fig.~\ref{fig:foot_geom}).
We compare two contact models.

\paragraph{Model A (shank-extension approximation).}
A common point-foot approximation is to treat the end-effector radius as a rigid extension of the shank and to assume a fixed stance contact point.
Let the (fictitious) contact point at touchdown be
\begin{equation}
\label{eq:foot_F0}
F_0 = O_1 + R
\begin{bmatrix}
\cos a_1\\
-\sin a_1
\end{bmatrix}.
\end{equation}
If the shank rotates to $a_2$ while keeping $F_0$ fixed, the implied connection point becomes
\begin{equation}
\label{eq:foot_O3}
O_3 = F_0 + R
\begin{bmatrix}
-\cos a_2\\
\sin a_2
\end{bmatrix}.
\end{equation}

\paragraph{Model B (hemispherical rolling contact).}
Under no-slip rolling, the touchdown contact is the vertical projection of $O_1$:
\begin{equation}
\label{eq:foot_F1}
F_1 = O_1 +
\begin{bmatrix}
0\\
-R
\end{bmatrix}.
\end{equation}
During stance, the contact point translates along the ground by the arc-length
\begin{equation}
\label{eq:foot_roll}
s = R\,(a_2-a_1),
\end{equation}
so the lift-off contact is $F_2 = F_1 + [s,\,0]^T$, and the corresponding connection point is
\begin{equation}
\label{eq:foot_O2}
O_2 = F_2 +
\begin{bmatrix}
0\\
R
\end{bmatrix}.
\end{equation}

\begin{figure}[htbp]
    \centering
    \includegraphics[width=0.6\linewidth]{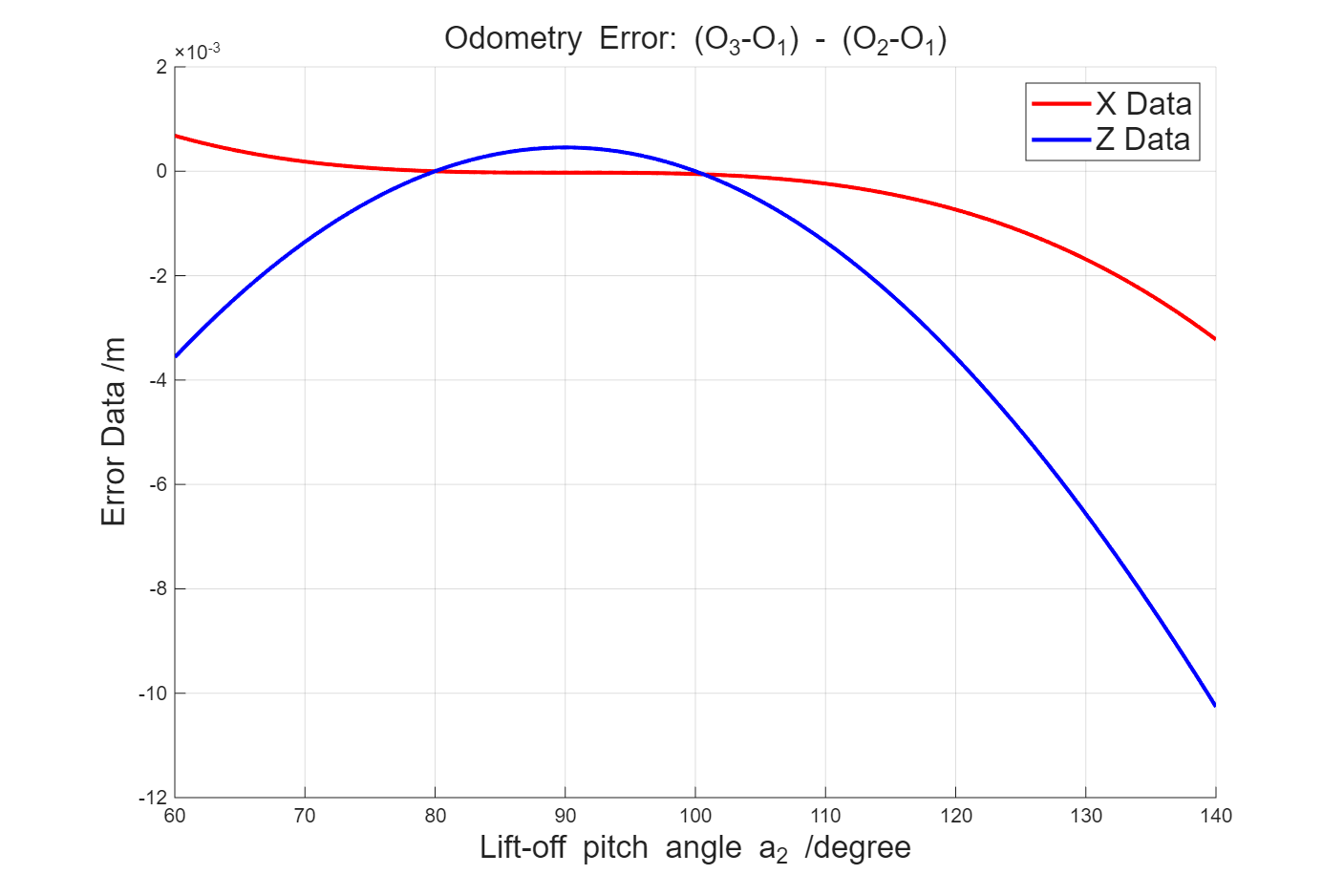}
    \caption{Rolling-contact modeling bias for $R=0.03$\,m and touchdown pitch $a_1=80^\circ$, with $a_2\in[60^\circ,140^\circ]$. The curves plot $\Delta_x$ (red) and $\Delta_z$ (blue) from \eqref{eq:foot_delta}.}
    \label{fig:foot_error}
\end{figure}

\paragraph{Displacement bias.}
We define the modeling error as the difference between the displacement predicted by Model~A and the physical displacement under Model~B:
\begin{equation}
\label{eq:foot_delta_def}
\Delta \triangleq (O_3-O_1)-(O_2-O_1)=O_3-O_2 =
\begin{bmatrix}\Delta_x\\ \Delta_z\end{bmatrix}.
\end{equation}
Substituting \eqref{eq:foot_F0}--\eqref{eq:foot_O2} yields the closed form
\begin{equation}
\label{eq:foot_delta}
\Delta_x = R\!\left(\cos a_1-\cos a_2-(a_2-a_1)\right),\qquad
\Delta_z = R\!\left(-\sin a_1+\sin a_2\right).
\end{equation}
The bias scales linearly with $R$ and grows with the stance pitch excursion $(a_2-a_1)$.

\paragraph{Implication for odometry.}
In contact-anchored estimation, the stance contact is modeled as stationary in the world frame.
When a rounded foot undergoes rolling while the estimator assumes a fixed contact point, the displacement bias given by \eqref{eq:foot_delta} manifests as a perturbation of the contact anchor.
However, the magnitude of this bias is bounded and directly proportional to the foot radius $R$.

Figure~\ref{fig:foot_error} reports the bias \eqref{eq:foot_delta} for a representative touchdown pitch $a_1=80^\circ$ and lift-off pitch $a_2\in[60^\circ,140^\circ]$ with $R=0.03$\,m.
While the bias increases toward larger $a_2$, the shank pitch during typical stance-to-swing transitions in our experiments is mostly within $a_2\in[60^\circ,120^\circ]$.
Over this practical range, the error remains below $\sim$5\,mm in $\Delta_z$ and $\sim$1\,mm in $\Delta_x$, which is small compared with dominant proprioceptive error sources (timing mismatch, encoder quantization, noise, slip, and compliance).
Accordingly, we model the rounded point foot as an effective shank extension to keep a unified and lightweight kinematic estimator.

Therefore, to preserve a unified kinematic formulation across platforms and to avoid introducing additional model complexity and state variables, we deliberately approximate the point foot as an effective shank extension.
This approximation yields negligible degradation in practical estimation accuracy while maintaining algorithmic consistency and computational simplicity.

\subsection{Wheel-Contact Propagation for Wheel-Legged Platforms}
\label{sec:wheel_comp}

For wheel-legged robots, the stance end-effector is not stationary even under no-slip contact: the ground contact point translates as the wheel rolls.
Therefore, the contact-anchored formulation of Section~\ref{sec:contact_anchored} must be augmented by a kinematic propagation of the recorded contact point.
In our implementation, this is achieved by estimating an \emph{effective wheel rolling angle} that removes the apparent rotation induced by shank pitching, by compensating lateral motion induced by shank roll, and then by integrating the corresponding displacement into the footfall record along a terrain-aware rolling direction.

\paragraph{Effective rolling angle.}
A wheel-legged robot may experience apparent wheel rotation even when the ground contact does not translate.
Consider the common stance scenario where the wheel is firmly constrained on the ground (no-slip) while the leg swings relative to the trunk during stepping.
In this case, the wheel--ground contact point and the wheel center remain approximately fixed in the world frame, yet the wheel motor encoder can still report a nonzero change in wheel angle and angular velocity.
This apparent rotation is induced by the shank pitch motion: as the shank pitches, the wheel joint axis rotates with respect to the world, producing encoder motion that should not be interpreted as true rolling along the ground.

To isolate the physically meaningful rolling component, we subtract the shank pitch contribution from the wheel encoder increment.
Let $\psi_{i,k}$ denote the wheel joint angle of leg $i$ at time $t_k$ (encoder) and $\dot{\psi}_{i,k}$ its angular velocity.
We define the shank pitch angle in the world frame as
\begin{equation}
\label{eq:shank_pitch}
\beta_{i,k} = \theta_{k}-\sum_{j\in\mathcal{P}_i} q_{i,j,k},
\end{equation}
where $\theta_k$ is the estimated body pitch and $\mathcal{P}_i$ contains the pitch joints whose motion rotates the wheel support link.
This notation includes the common thigh--calf compensation as a special case, while matching the transform-chain implementation where the compensation joints are configured per robot.
The wheel encoder increment is wrapped to avoid discontinuities,
\begin{equation}
\label{eq:wheel_wrap}
\Delta\psi_{i,k} = \mathrm{wrap}(\psi_{i,k}-\psi_{i,k-1})\in[-\pi,\pi],
\end{equation}
and the shank pitch increment is $\Delta\beta_{i,k}=\beta_{i,k}-\beta_{i,k-1}$.
The effective rolling increment is then computed as
\begin{equation}
\label{eq:wheel_eff_angle}
\Delta\psi^{\mathrm{eff}}_{i,k} = \Delta\psi_{i,k} - \Delta\beta_{i,k},
\end{equation}
which approximates the net wheel rotation relative to the ground during stance by removing the pitch-induced encoder motion.

\paragraph{Heading direction on the ground plane.}
Let $\mathbf{e}_x=[1,0,0]^\top$ be the body forward axis.
We compute its world-frame direction $\mathbf{h}_k=\mathbf{R}_{WB,k}\mathbf{e}_x$ and project to the horizontal plane:
\begin{equation}
\label{eq:heading_proj}
\hat{\mathbf{t}}_k
=
\frac{1}{\sqrt{h_{x,k}^2+h_{y,k}^2}}
\begin{bmatrix}
h_{x,k}\\
h_{y,k}\\
0
\end{bmatrix},
\qquad
\text{if } \sqrt{h_{x,k}^2+h_{y,k}^2}>\varepsilon_h,
\end{equation}
where $\varepsilon_h$ is a small numerical threshold (in code: $10^{-9}$).
This matches the implementation that extracts $(h_x,h_y)$ directly from the current attitude quaternion and normalizes the planar component.
When slope-aware propagation is enabled, reliable contacts are additionally used to fit a local support plane, $z=ax+by+c$, and the rolling direction is lifted to the tangent of this plane along the current heading. If the fitted slope is below the angular threshold or the contact set is insufficient, the horizontal direction above is retained.

\paragraph{Contact-point propagation.}
Let $\mathbf{c}^W_{i,k}$ denote the world-frame contact (footfall) record of leg $i$.
For wheel-legged stance, the contact record is no longer assumed to be fixed.
Let $\hat{\mathbf{d}}_k=[d_{x,k},d_{y,k},d_{z,k}]^T$ denote the rolling direction, which reduces to the horizontal heading on flat terrain and may contain a vertical component under slope-aware propagation.
The contact record is propagated by
\begin{equation}
\label{eq:contact_propagation}
\mathbf{c}^W_{i,k}
\leftarrow
\mathbf{c}^W_{i,k-1}
+
r_w\,\Delta\psi^{\mathrm{eff}}_{i,k}\,\hat{\mathbf{d}}_k
+
\Delta s^{\mathrm{side}}_{i,k}
\begin{bmatrix}
-d_{y,k}\\
d_{x,k}\\
0
\end{bmatrix}.
\end{equation}
Here $r_w$ is the wheel radius (parameter $\texttt{Par\_WheelRadius}$).
In addition, shank roll induces a small lateral displacement of the wheel contact. With shank roll angle $\gamma_{i,k}$, the sideway term is approximated by
\begin{equation}
\label{eq:wheel_sideway}
\Delta s^{\mathrm{side}}_{i,k}=2r_w\sin\left(\frac{\mathrm{wrap}(\gamma_{i,k}-\gamma_{i,k-1})}{2}\right),
\end{equation}
and is integrated along the direction perpendicular to the rolling direction.
This step is applied only when the leg is deemed in contact, and it reduces systematic bias that would arise if a rolling wheel were treated as a fixed stance anchor.

\paragraph{Velocity augmentation.}
The same idea is used to augment the stance-based velocity observation.
In implementation we subtract the configured joint-induced shank pitch rate.
Using the encoder angular velocity and joint angular velocities, we form an effective wheel rolling rate
\begin{equation}
\label{eq:wheel_eff_rate}
\dot\psi^{\mathrm{eff}}_{i,k} = \dot\psi_{i,k} - \sum_{j\in\mathcal{P}_i}\dot q_{i,j,k},
\end{equation}
consistent with the implementation (wheel rate minus the shank pitch rate induced by the leg joints).
The induced rolling velocity is
\begin{equation}
\label{eq:wheel_vel}
\mathbf{v}^{W}_{\mathrm{roll},i,k} = r_w\,\dot\psi^{\mathrm{eff}}_{i,k}\,\hat{\mathbf{d}}_k,
\end{equation}
which is added to the kinematics-derived stance velocity term when forming the fused body velocity observation (cf.~\eqref{eq:fuse_obs2}).

\paragraph{Degenerate case and point-foot platforms.}
If $r_w=0$, the propagation \eqref{eq:contact_propagation}--\eqref{eq:wheel_vel} vanishes and the formulation reduces to the point-foot case where the stance anchor is treated as stationary.
In our parameterization, point-foot robots absorb the foot radius into the effective shank length (Section~III-A), whereas wheel-legged robots keep $r_w>0$ and apply the explicit rolling propagation above.

\section{Inverse-Kinematics Cubature Kalman Filtering for Foot-End Velocity}
\label{sec:ikvel_ckf}

Contact-anchored proprioceptive odometry (Section~\ref{sec:contact_anchored}) relies on stance-phase kinematics to provide body-velocity observations.
A practical challenge is that joint angular velocities obtained from motor encoders are affected by quantization and discrete-time differentiation, and their noise is amplified by the nonlinear leg kinematics.
As a result, the derived hip-to-foot velocity can exhibit impulsive spikes, which then propagate into the fused body velocity and degrade the stability of contact-anchored updates.
To suppress these artifacts while keeping the pipeline purely proprioceptive, we retain IKVel-CKF as the CAPO-CKE enhancement: it estimates the hip-to-end-effector Cartesian velocity directly from joint angles and joint velocities and can replace the default Jacobian velocity when smoother velocity feedback is required.
The analytic inverse-kinematics mapping and the CKF recursion used in our implementation are provided in Appendix~\ref{app:ikvel_ckf}.

\subsection{Motivation: Spurious Velocity Spikes from Encoder-Derived Joint Rates}
\label{sec:ikvel_motivation}

Figure~\ref{fig:vel_body} (left) shows the estimated body linear velocity when the system uses raw encoder-derived joint rates without IKVel-CKF.
The horizontal channels ($v_x$ and $v_y$) contain pronounced spikes, reaching up to roughly $200\%$ of the nominal signal magnitude.
These spikes are consistent with encoder quantization, discrete differentiation, and slight timing mismatch, which are then magnified by nonlinear kinematic projection.
The issue is further illustrated at the leg level in Fig.~\ref{fig:legvel}, where the raw hip--foot velocity of a representative leg (blue) contains impulsive excursions.

\subsection{IKVel-CKF: Nonlinear Filtering under an Inverse-Kinematics Measurement Model}
\label{sec:ikvel_method}

For each leg $i$, IKVel-CKF maintains a 6D latent Cartesian state
\begin{equation}
\label{eq:ikvel_state_main}
\mathbf{x}_{i,k} \triangleq
\begin{bmatrix}
\mathbf{r}_{i,k} \\
\dot{\mathbf{r}}_{i,k}
\end{bmatrix}
\in \mathbb{R}^6,
\end{equation}
where $\mathbf{r}_{i,k}$ and $\dot{\mathbf{r}}_{i,k}$ are the hip-to-end-effector relative position and velocity.
A constant-velocity prior propagates $\mathbf{x}_{i,k}$ between timestamps, while the measurement is the joint configuration and joint velocity,
\begin{equation}
\label{eq:ikvel_meas_main}
\mathbf{z}_{i,k} \triangleq
\begin{bmatrix}
\mathbf{q}_{i,k} \\
\dot{\mathbf{q}}_{i,k}
\end{bmatrix}
=
\mathbf{h}(\mathbf{x}_{i,k}) + \mathbf{v}_{i,k},
\end{equation}
where $\mathbf{h}(\cdot)$ is an analytic inverse-kinematics and differential inverse-kinematics mapping.
Because $\mathbf{h}(\cdot)$ is strongly nonlinear and may be piecewise-defined due to kinematic branches, we employ a CKF to update $\mathbf{x}_{i,k}$ without requiring Jacobians.
In implementation, each leg caches $(\mathbf{x}_i,\mathbf{P}_i,t_i)$, and a single CKF instance is reused by swapping in/out the cached state for efficiency.
The filtered Cartesian velocity $\dot{\mathbf{r}}_{i,k}$ then replaces the raw forward-kinematics velocity before transforming to the world frame and contributing to stance velocity constraints.

\subsection{Effect on Body Velocity Estimation}
\label{sec:ikvel_effect}

\begin{figure*}[htbp]
    \centering
    \begin{minipage}[b]{0.48\linewidth}
        \centering
        \includegraphics[width=\linewidth]{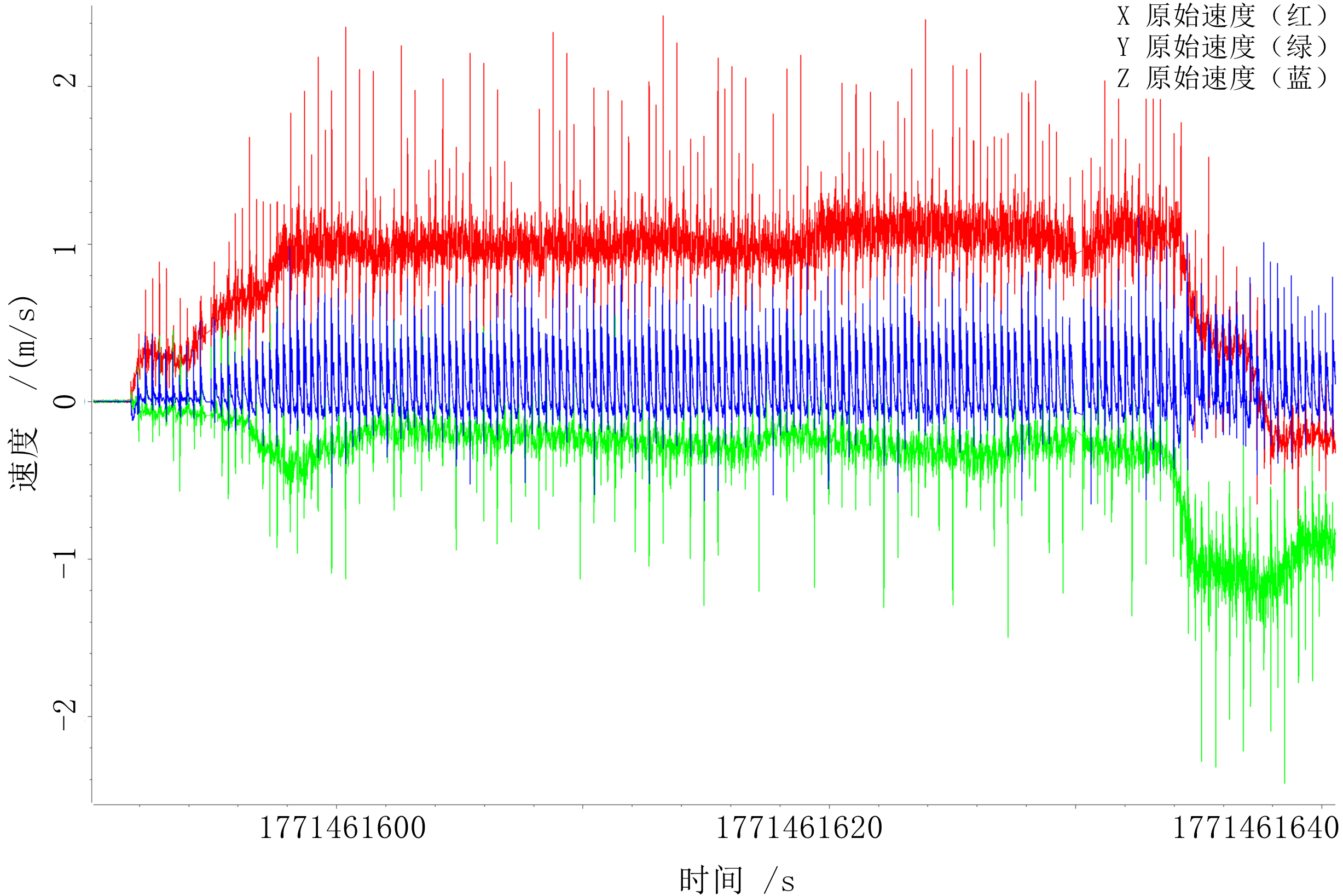}\\[-0.6ex]
        {\footnotesize (a) Without IKVel-CKF.}
    \end{minipage}
    \hfill
    \begin{minipage}[b]{0.48\linewidth}
        \centering
        \includegraphics[width=\linewidth]{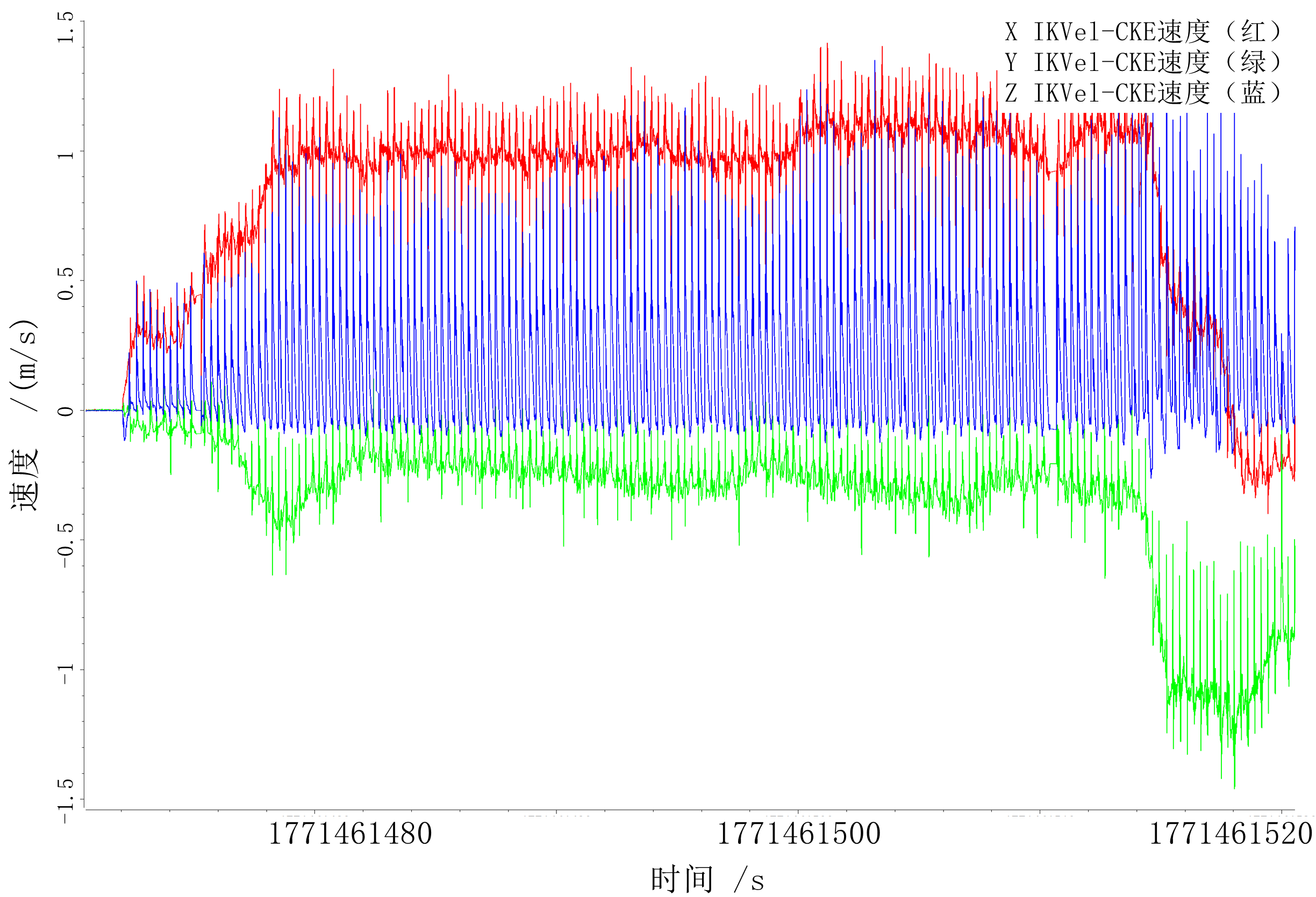}\\[-0.6ex]
        {\footnotesize (b) With IKVel-CKF.}
    \end{minipage}
    \caption{Estimated body linear velocity. Red/green/blue correspond to $v_x$, $v_y$, and $v_z$. Without IKVel-CKF, the horizontal channels exhibit large impulsive errors (up to $\sim$200\% of nominal magnitude). With IKVel-CKF enabled, the spikes are strongly suppressed (within $\sim$25\% of nominal magnitude).}
    \label{fig:vel_body}
\end{figure*}

\begin{figure*}[htbp]
    \centering
    \begin{minipage}[b]{0.48\linewidth}
        \centering
        \includegraphics[width=\linewidth]{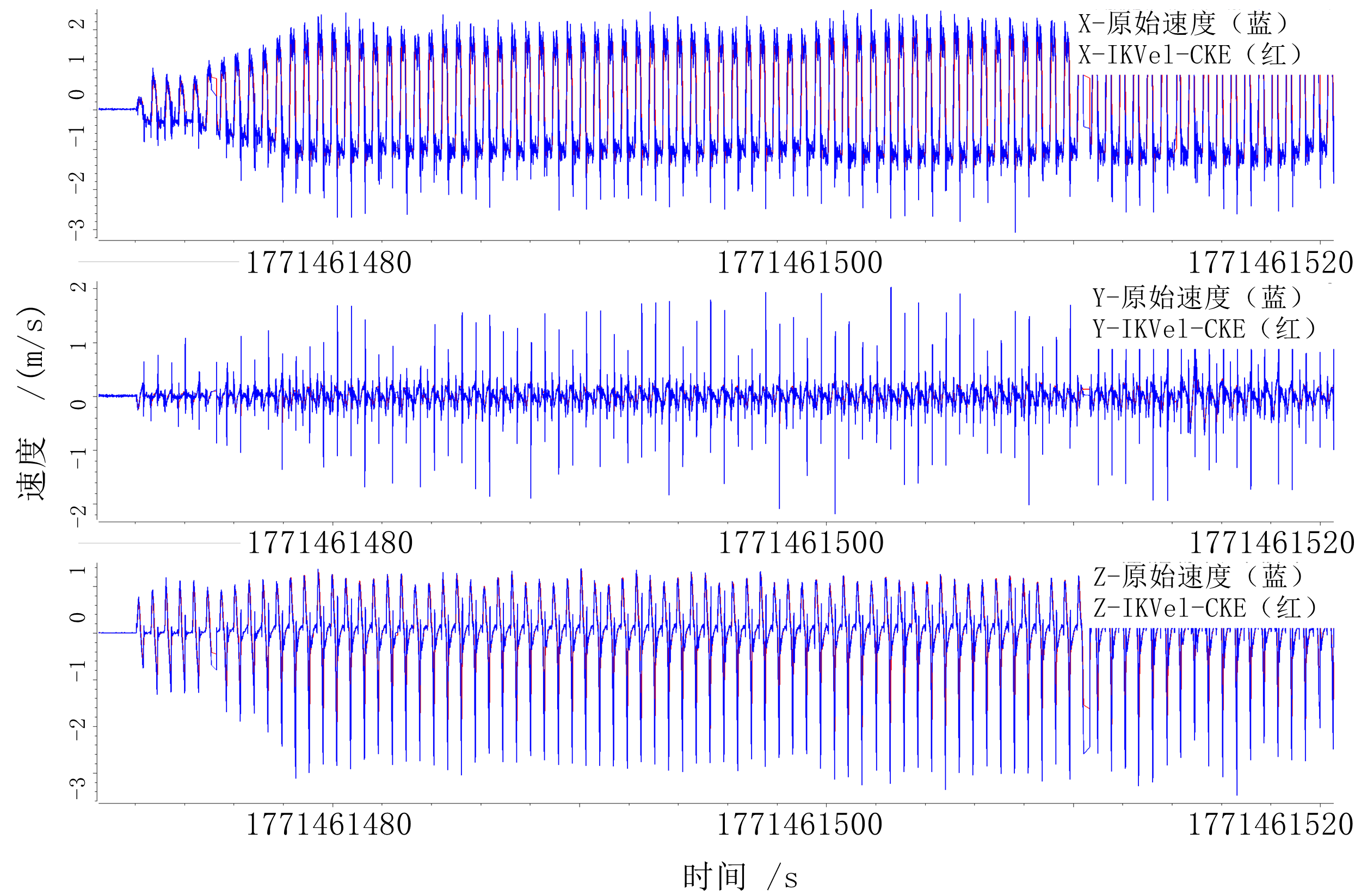}\\[-0.6ex]
        {\footnotesize (a) Overview.}
    \end{minipage}
    \hfill
    \begin{minipage}[b]{0.48\linewidth}
        \centering
        \includegraphics[width=\linewidth]{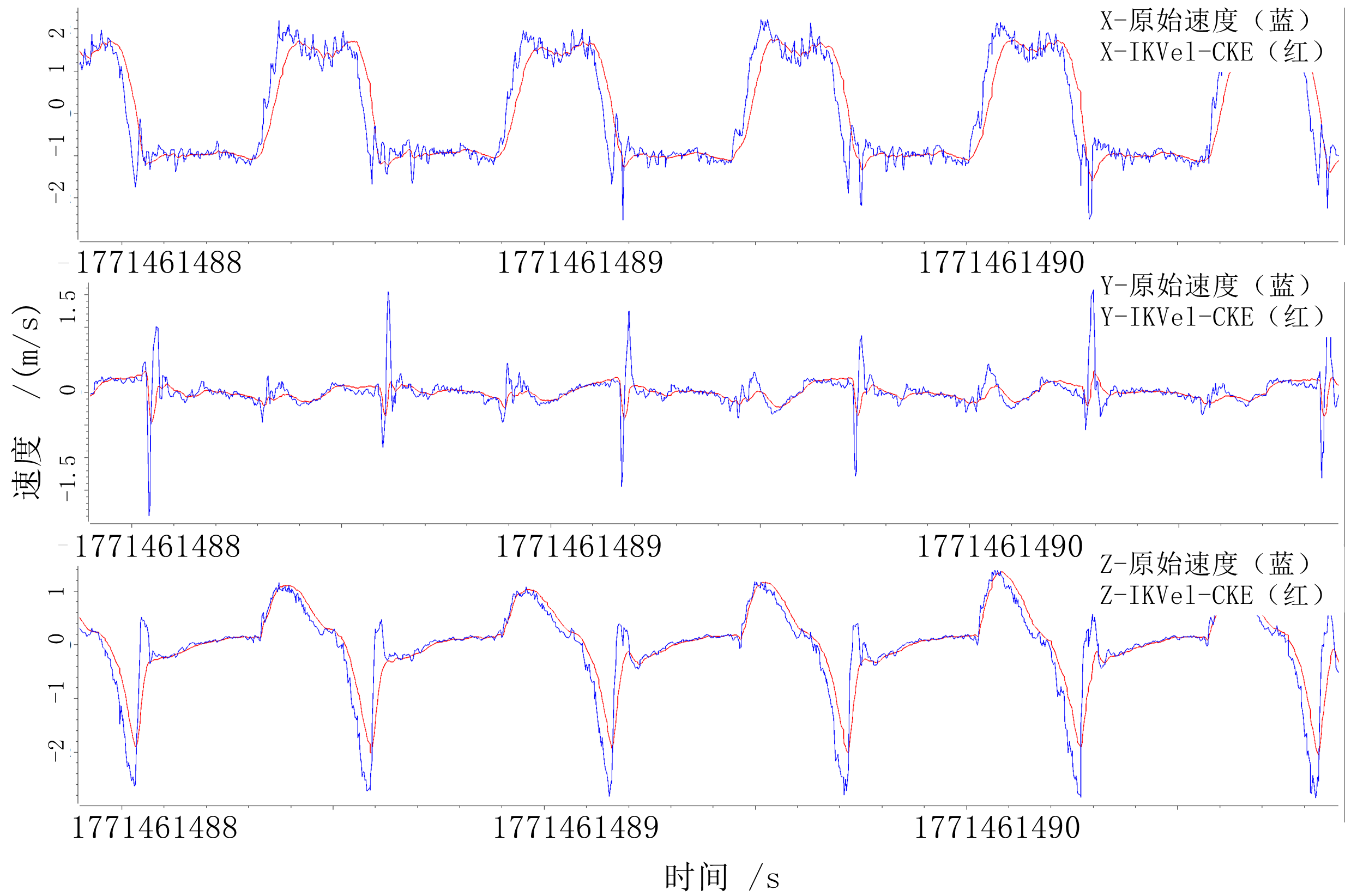}\\[-0.6ex]
        {\footnotesize (b) Zoomed view.}
    \end{minipage}
    \caption{Representative hip--foot velocity feedback (leg~1). Blue: raw forward-kinematics velocity using encoder-derived joint rates. Red: IKVel-CKF filtered velocity. IKVel-CKF effectively removes impulsive spikes; the zoomed view shows a modest smoothing-induced lag, trading a small phase delay for substantially improved robustness.}
    \label{fig:legvel}
\end{figure*}

After enabling IKVel-CKF, the estimated body velocity becomes substantially more stable.
As shown in Fig.~\ref{fig:vel_body} (right), the horizontal velocity spikes are reduced to within roughly $25\%$ of the nominal magnitude.
At the leg level (Fig.~\ref{fig:legvel}), IKVel-CKF (red) effectively suppresses impulsive excursions compared with the raw signal (blue).
A zoomed view (Fig.~\ref{fig:legvel}, right) indicates a modest smoothing-induced lag, but the reduction of high-frequency noise and spike removal yields a more reliable velocity feedback for contact-anchored fusion.

In the CAPO-CKE variant, the IKVel-CKF output replaces the raw encoder-derived hip--foot velocity in the stance-phase velocity constraint of Section~\ref{sec:contact_anchored}, so that the contact-anchored update uses a denoised end-effector velocity observation.
The default released estimator, by contrast, uses the transform-chain Jacobian velocity directly for lightweight real-time operation.

\paragraph{Computation--accuracy trade-off.}
In our MATLAB/MEX deployment, enabling IKVel-CKF increases the per-cycle computation time by more than a factor of two due to the nonlinear cubature updates.
Accordingly, IKVel-CKF is best treated as an optional module: when reliable outer-loop contact-anchored position feedback is available from footfall constraints, the impact of IKVel-CKF on overall position accuracy is limited, while its primary benefit remains improved velocity smoothness and robustness; thus it should be enabled only when required by the downstream control or estimation objectives.

\section{Kinematics-Based Yaw Estimation via Multi-Contact Geometric Consistency}
\label{sec:yaw_kin}

Yaw is weakly observable from inertial sensing alone because gravity does not constrain heading.
As a result, IMU-based yaw can drift during long-horizon operation, and the effect becomes especially visible during prolonged standing, where the robot should remain stationary but the integrated yaw slowly wanders.
This section presents a proprioceptive yaw correction method that leverages the \emph{geometric consistency} between (i) the current body-frame end-effector configuration computed from joint kinematics and (ii) the world-frame contact (footfall) records maintained by the contact-anchored odometry in Section~\ref{sec:contact_anchored}.
The key idea is simple: when multiple contacts are simultaneously stationary, the relative geometry between contact points is fixed in the world, and any discrepancy between this world geometry and the tilt-compensated body geometry reveals the yaw misalignment.

\begin{figure}[t]
    \centering
    \begin{minipage}[b]{0.48\linewidth}
        \centering
        \includegraphics[width=\linewidth]{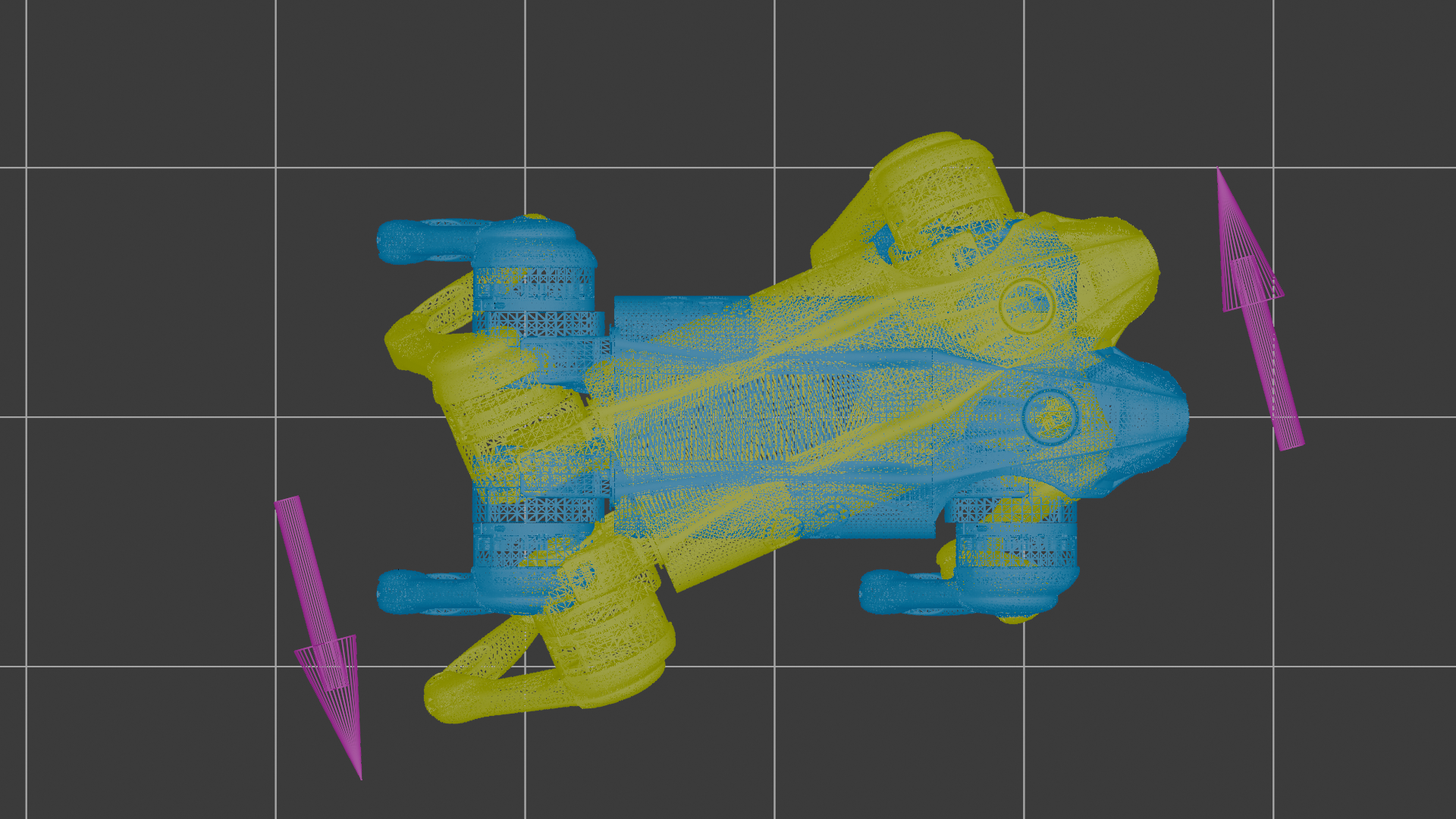}\\[-0.6ex]
        {\footnotesize (a) Top view (yaw twist with fixed contacts).}
    \end{minipage}
    \hfill
    \begin{minipage}[b]{0.48\linewidth}
        \centering
        \includegraphics[width=\linewidth]{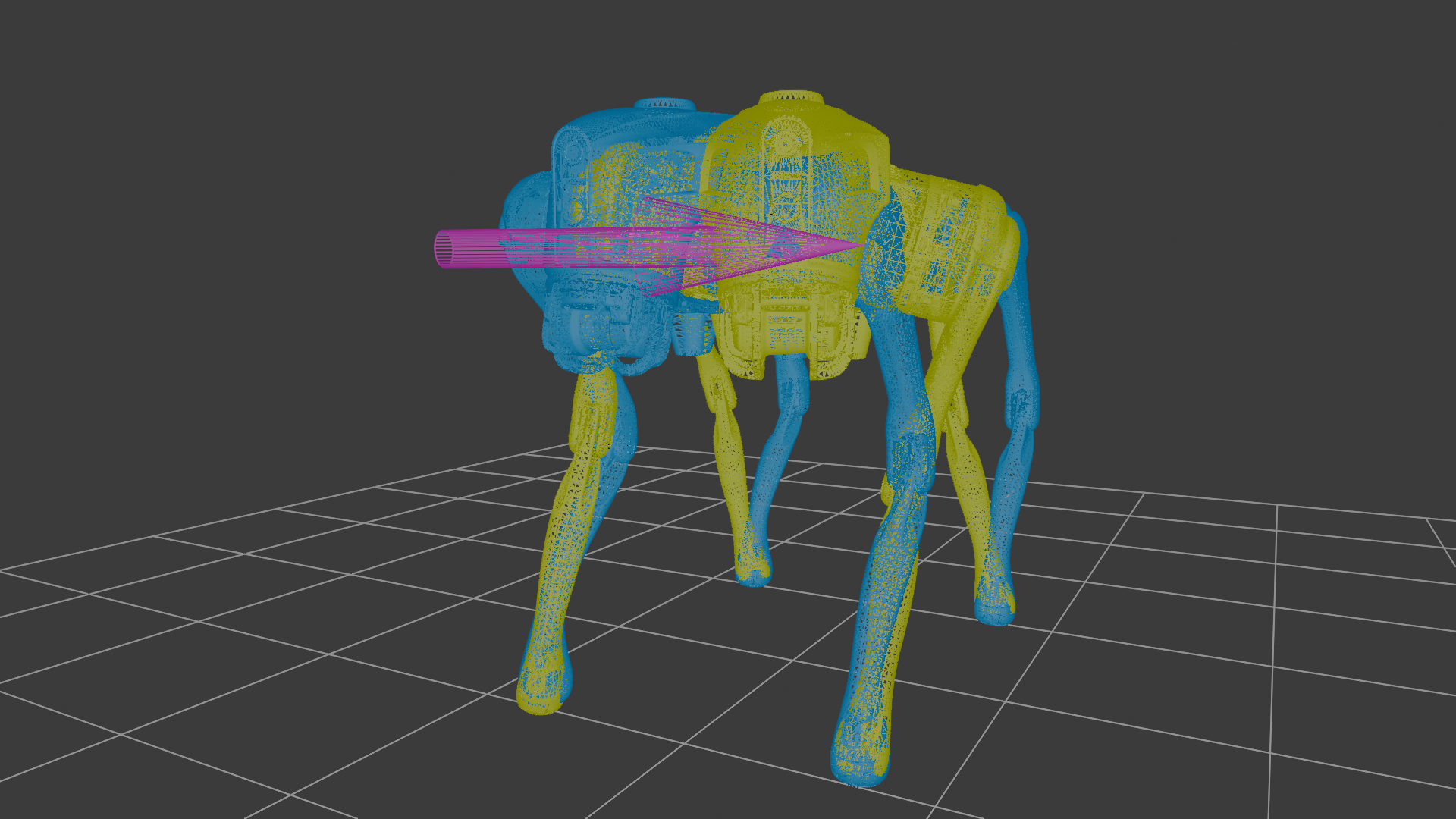}\\[-0.6ex]
        {\footnotesize (b) Front view.}
    \end{minipage}
    \caption{Illustration of the geometric cue exploited for yaw correction. The blue model denotes the initial configuration and the yellow model denotes the twisted trunk pose while feet remain fixed. With multi-contact, the world-frame inter-foot geometry stays constant, enabling a kinematics-derived heading reference.}
    \label{fig:yaw_rotation}
\end{figure}

\subsection{Pairwise Yaw from Contact Geometry}
\label{sec:yaw_pairwise}

Let $\mathcal{C}_k$ denote the set of end-effectors in stance at time $t_k$ (Section~\ref{sec:sec2_stance}).
For each contacting leg $i\in\mathcal{C}_k$, we use:
(1) the body-frame end-effector position $\mathbf{p}^{B}_{\mathrm{ee},i,k}$ from forward kinematics; and
(2) the stored world-frame footfall record $\mathbf{c}^{W}_i$ (assumed stationary during stance).
To eliminate dependence on global translation and improve robustness, we form pairwise relative vectors:
\begin{equation}
\label{eq:yaw_relvec}
\mathbf{v}^{B}_{ij,k} = \mathbf{p}^{B}_{\mathrm{ee},j,k} - \mathbf{p}^{B}_{\mathrm{ee},i,k}, \qquad
\mathbf{v}^{W}_{ij}   = \mathbf{c}^{W}_{j} - \mathbf{c}^{W}_{i},
\end{equation}
for all unordered pairs $(i,j)$ with $i,j\in\mathcal{C}_k$ and $i<j$.

We assume roll and pitch are available from the state estimator (e.g., IMU tilt) and isolate heading by constructing a \emph{tilt-only} rotation
\begin{equation}
\label{eq:yaw_rp}
\mathbf{R}_{rp,k} \triangleq \mathbf{R}_{y}(\theta_k)\mathbf{R}_{x}(\phi_k),
\end{equation}
where $(\phi_k,\theta_k)$ are the estimated roll and pitch.
We then tilt-compensate the body-frame relative vector:
\begin{equation}
\label{eq:yaw_tiltcomp}
\bar{\mathbf{v}}_{ij,k} = \mathbf{R}_{rp,k}\mathbf{v}^{B}_{ij,k}.
\end{equation}
The yaw implied by pair $(i,j)$ is computed from the difference between the horizontal bearings of $\mathbf{v}^{W}_{ij}$ and $\bar{\mathbf{v}}_{ij,k}$:
\begin{equation}
\label{eq:yaw_ij}
\hat{\psi}_{ij,k} =
\mathrm{wrap}\!\left(
\mathrm{atan2}\big((\mathbf{v}^{W}_{ij})_y,(\mathbf{v}^{W}_{ij})_x\big)
-
\mathrm{atan2}\big((\bar{\mathbf{v}}_{ij,k})_y,(\bar{\mathbf{v}}_{ij,k})_x\big)
\right),
\end{equation}
where $\mathrm{wrap}(\cdot)$ maps angles to $(-\pi,\pi]$ to avoid discontinuities.
Finally, we aggregate all available pairs using a circular mean. In the implementation, each pair is weighted by the product of the two soft contact confidences, $w_{ij,k}=\rho_{i,k}\rho_{j,k}$, so weak or uncertain contacts contribute less to the yaw cue:
\begin{equation}
\label{eq:yaw_circmean}
\hat{\psi}_k =
\mathrm{atan2}\!\left(
\sum_{(i,j)} w_{ij,k}\sin(\hat{\psi}_{ij,k}),
\sum_{(i,j)} w_{ij,k}\cos(\hat{\psi}_{ij,k})
\right).
\end{equation}

The estimator requires at least two simultaneous contacts ($|\mathcal{C}_k|\ge 2$); otherwise no geometric yaw constraint is applied.
Using multiple pairs increases robustness by averaging over baselines of different lengths and orientations, and it naturally rejects wrap-around artifacts via \eqref{eq:yaw_circmean}.

\subsection{Drift Suppression and Kinematics-Only Fallback}
\label{sec:yaw_filter}

We use the kinematics-derived heading $\hat{\psi}_k$ as a stabilizing reference for the yaw observation rather than as a hard reset of the attitude state.
Let
\begin{equation}
\label{eq:yaw_err}
e_k = \mathrm{wrap}(\hat{\psi}_k - \psi_k).
\end{equation}
The correction target is
\begin{equation}
\label{eq:yaw_update}
\psi^{\mathrm{corr}}_k = \mathrm{wrap}\!\left(\psi_k + \alpha_k e_k\right),
\end{equation}
and the corresponding yaw offset is injected into the subsequent yaw observation used by the attitude estimator.
Equivalently, if $z^{\psi}_{k,\mathrm{imu}}$ denotes the yaw observation from the IMU attitude channel, the corrected observation is
\begin{equation}
\label{eq:yaw_obs_bias}
z^{\psi}_{k}=z^{\psi}_{k,\mathrm{imu}}+\mathrm{wrap}(\psi^{\mathrm{corr}}_k-\psi_k).
\end{equation}
Thus the geometric cue acts as a soft heading prior while preserving the estimator's own temporal filtering.

Our implementation uses a contact-dependent schedule for $\alpha_k$ that matches the code logic.
When fewer than all configured end-effectors are in stance (typical during walking, or when contacts are uncertain), we reset the accumulation timer and keep a small gain $\alpha_k=\alpha_0$ to avoid over-correcting under intermittent contact changes.
When all configured contacts are simultaneously in stance and remain stable (prolonged standing), we ramp $\alpha_k$ linearly from $\alpha_0$ to $1$ over a time constant $T_\psi$:
\begin{equation}
\label{eq:yaw_gain}
\alpha_k =
\mathrm{clip}\!\left(
\alpha_0 + \frac{t_k-t_0}{T_\psi}(1-\alpha_0),\, 0,\, 1
\right),
\end{equation}
where $t_0$ is the time when full support is first detected.
This design has an important practical consequence: during long stationary standing with stable multi-contact, the yaw observation correction \eqref{eq:yaw_obs_bias} continuously re-anchors the heading to the contact geometry, effectively arresting IMU yaw drift.

When IMU yaw constraints are disabled, \eqref{eq:yaw_circmean}--\eqref{eq:yaw_update} still provide a purely kinematics-derived heading reference.
In this mode, the heading is no longer tied to inertial integration, but residual drift can still arise from unmodeled attitude coupling, contact compliance, and minor slip; therefore, we treat the kinematics-derived yaw primarily as a stabilizing prior rather than an absolute compass.

\section{Experimental Evaluation}
\label{sec:eval}

\subsection{Physics Simulation in Gazebo}
\label{sec:eval_sim}

We first evaluate the proposed estimator in physics simulation using Unitree's AlienGo model in Gazebo.
We compare (i) a Lidar-based SLAM baseline following \cite{DC2022}, (ii) \textbf{CAPO}, our contact-anchored proprioceptive odometry, and (iii) \textbf{CAPO-CKE}, which denotes CAPO with the IKVel-CKF module (Section~IV) enabled.
Gazebo ground-truth base states are used for error computation.

\begin{figure}[htbp]
    \centering
    \includegraphics[width=3in]{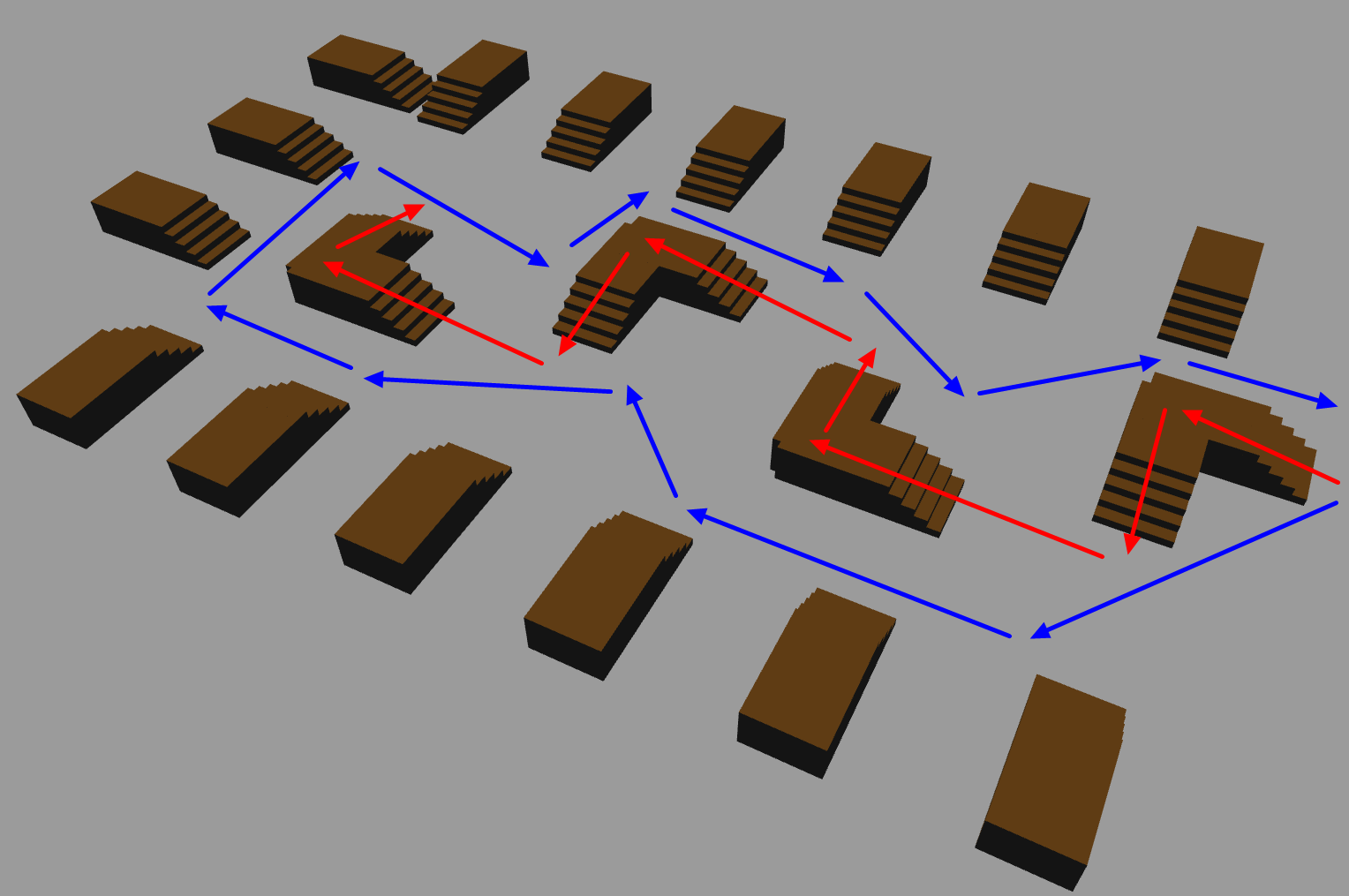}
    \caption{Locomotion trajectories in simulation. The flat-ground loop is shown in blue and the stair-climbing loop in red.}
    \label{P4_2_2}
\end{figure}

Two closed-loop trajectories are executed (Fig.~\ref{P4_2_2}): a flat-ground loop with obstacle avoidance (blue), and a stair-climbing loop (red).
All methods are evaluated on the same commanded motions.

\subsubsection{Flat-Terrain Loop}
\label{sec:eval_sim_flat}

\begin{figure*}[htbp]
    \centering
    \begin{minipage}[b]{0.32\linewidth}
        \centering
        \includegraphics[width=1\linewidth]{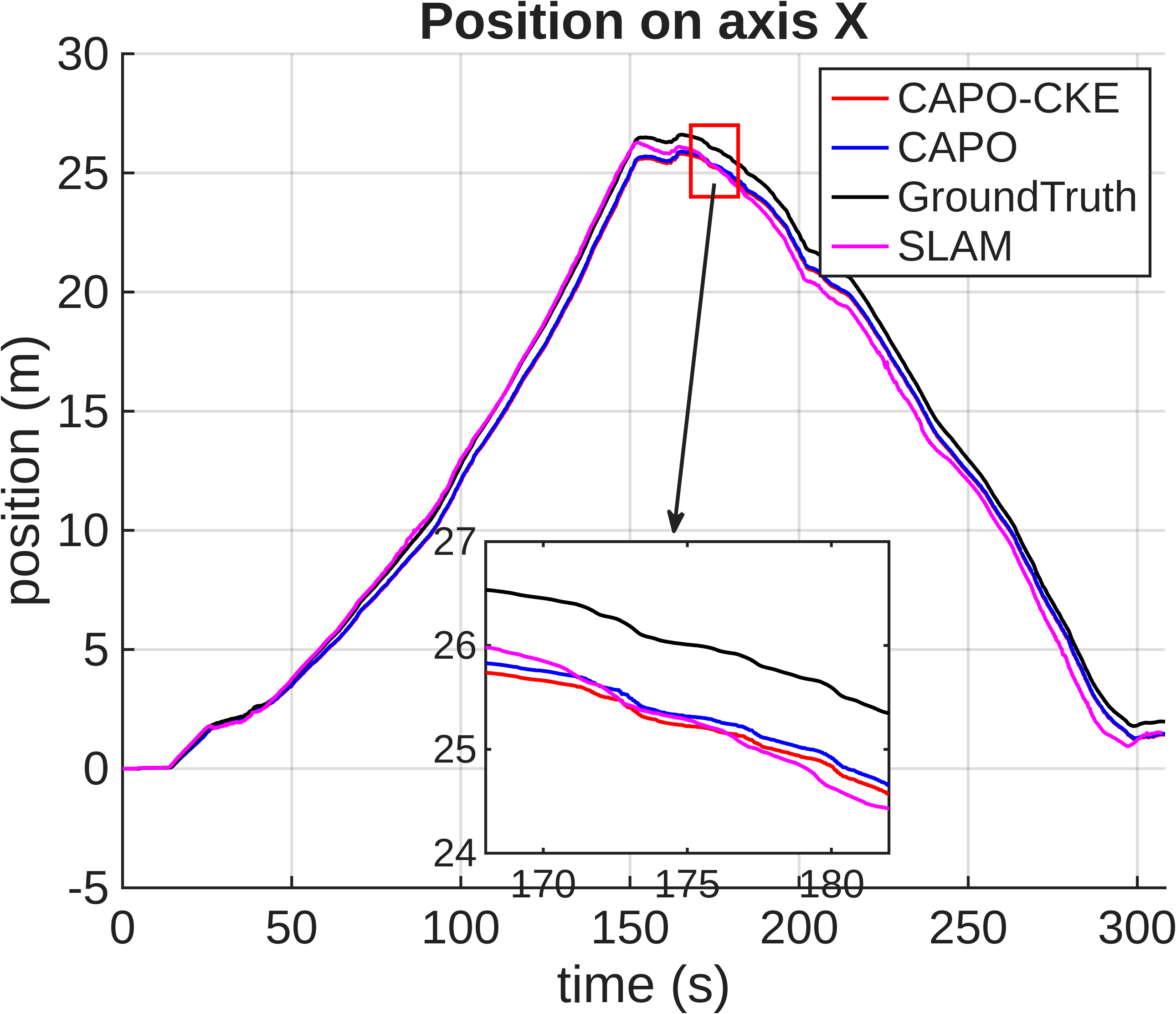}
        \caption{Estimated x-position during flat-ground walking.}
        \label{P4_2_3}
    \end{minipage}
    \hfill
    \begin{minipage}[b]{0.32\linewidth}
        \centering
        \includegraphics[width=1\linewidth]{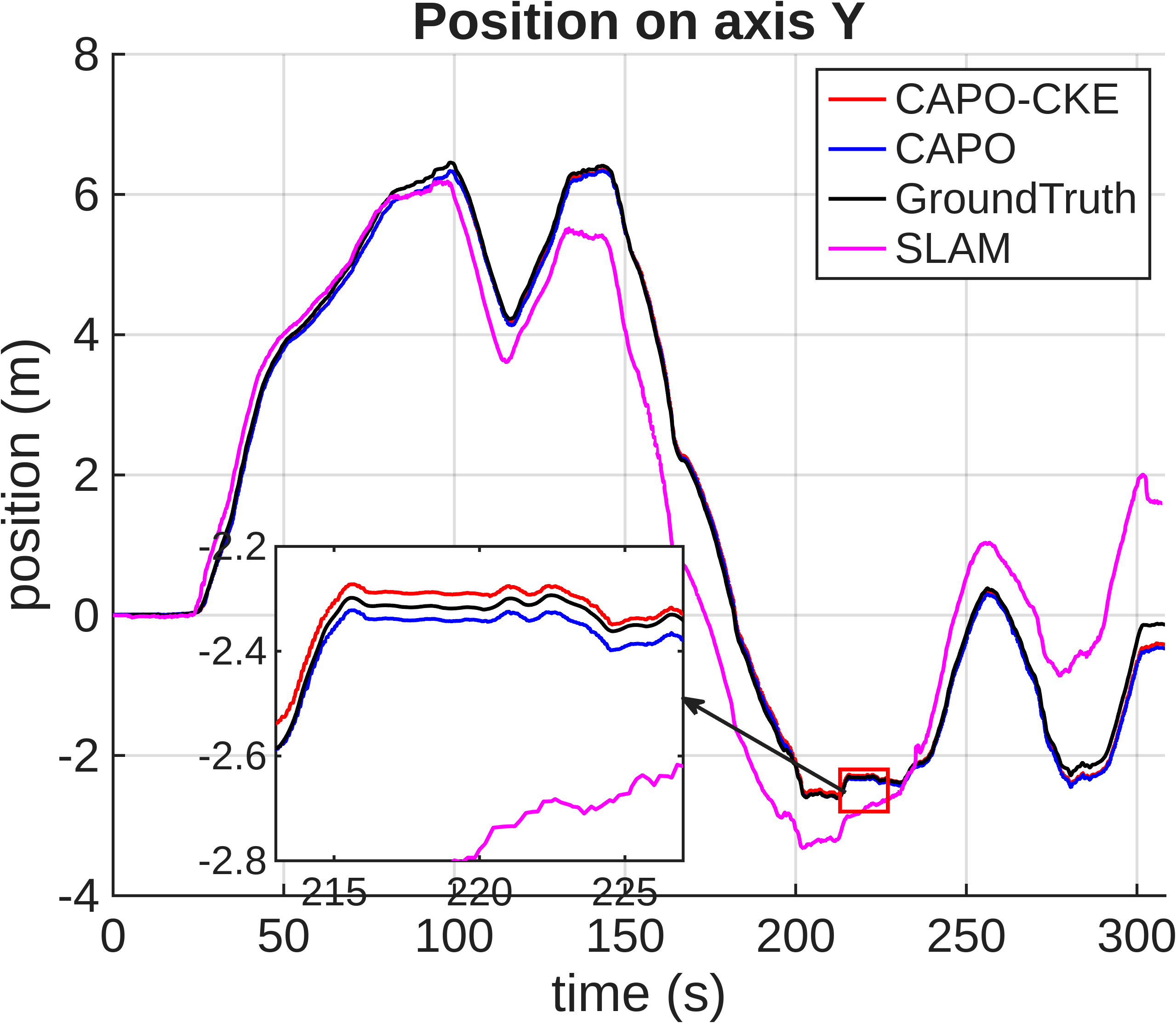}
        \caption{Estimated y-position during flat-ground walking.}
        \label{P4_2_4}
    \end{minipage}
    \hfill
    \begin{minipage}[b]{0.32\linewidth}
        \centering
        \includegraphics[width=1\linewidth]{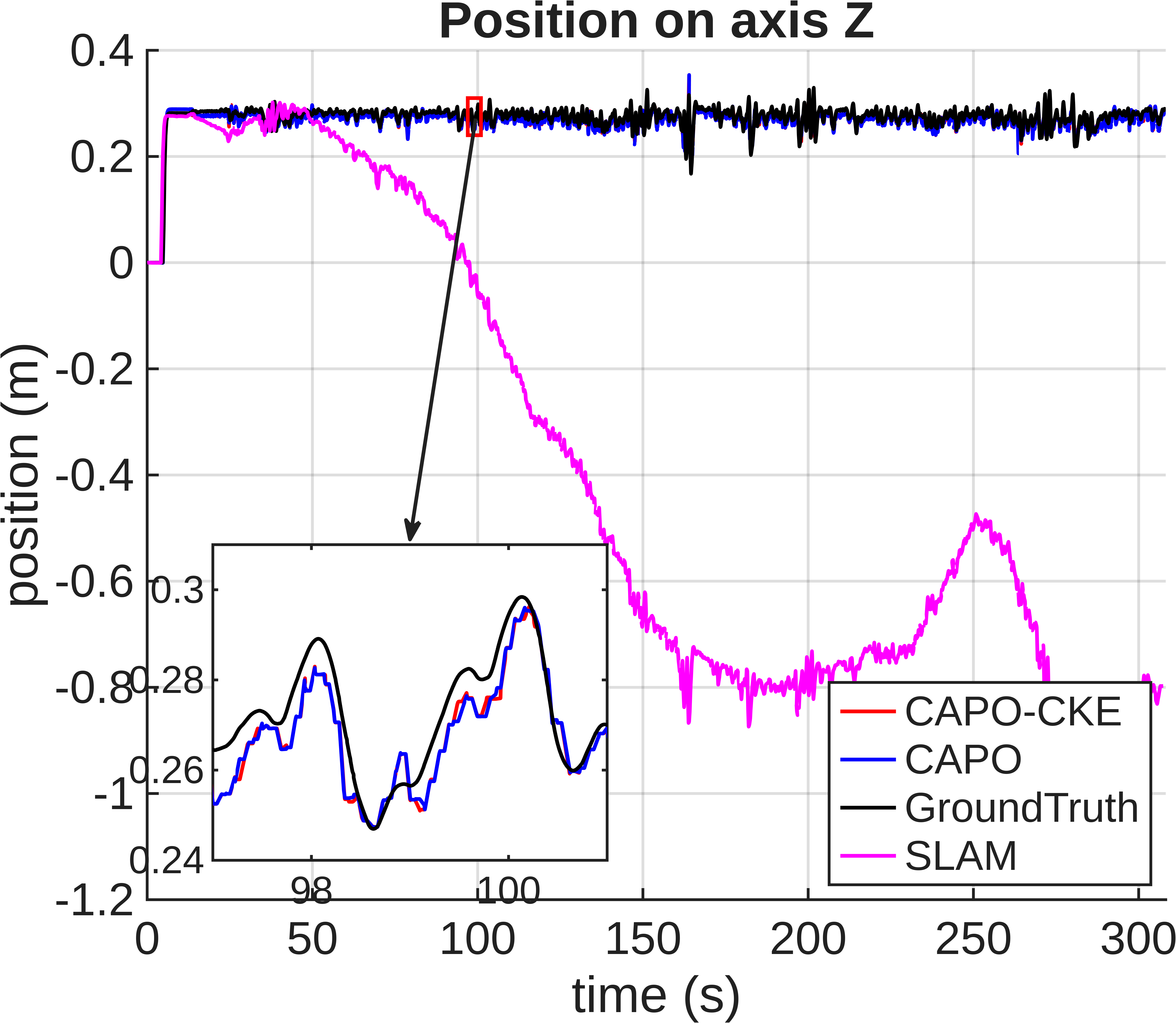}
        \caption{Estimated z-position during flat-ground walking.}
        \label{P4_2_5}
    \end{minipage}
\end{figure*}

Figure~\ref{P4_2_3}--\ref{P4_2_5} report the estimated base position components during the flat-ground loop.
Both proprioceptive methods (CAPO and CAPO-CKE) maintain stable elevation estimates and substantially reduce drift compared to the SLAM baseline, particularly along the vertical channel.
On this trajectory, CAPO and CAPO-CKE achieve comparable terminal accuracy; this is consistent with the fact that, under sustained stance constraints, the contact-anchored position update dominates long-horizon drift suppression, and additional velocity denoising mainly improves robustness under stronger excitation.

Table~\ref{T4_2_0} summarizes mean absolute errors (MAE) along each axis and the Euclidean terminal-position error.

\begin{table}[htbp]
    \centering
    \caption{Estimation error during flat-ground walking in simulation.}
    \label{T4_2_0}
    \begin{tabular}{|l|c|c|c|}
    \hline
    \textbf{Error} & \textbf{CAPO-CKE} & \textbf{CAPO} & \textbf{SLAM} \\
    \hline
    X MAE & 0.5671\,m & 0.5099\,m & 0.5329\,m \\
    Y MAE & 0.0818\,m & 0.0885\,m & 0.7149\,m \\
    Z MAE & 0.0072\,m & 0.0073\,m & 0.6503\,m \\
    Terminal error  & 0.6157\,m & 0.6232\,m & 2.078\,m \\
    \hline
    \end{tabular}
\end{table}

\subsubsection{Stair-Climbing Loop}
\label{sec:eval_sim_stairs}

\begin{figure*}[htbp]
    \centering
    \begin{minipage}[b]{0.48\linewidth}
        \centering
        \includegraphics[width=1\linewidth]{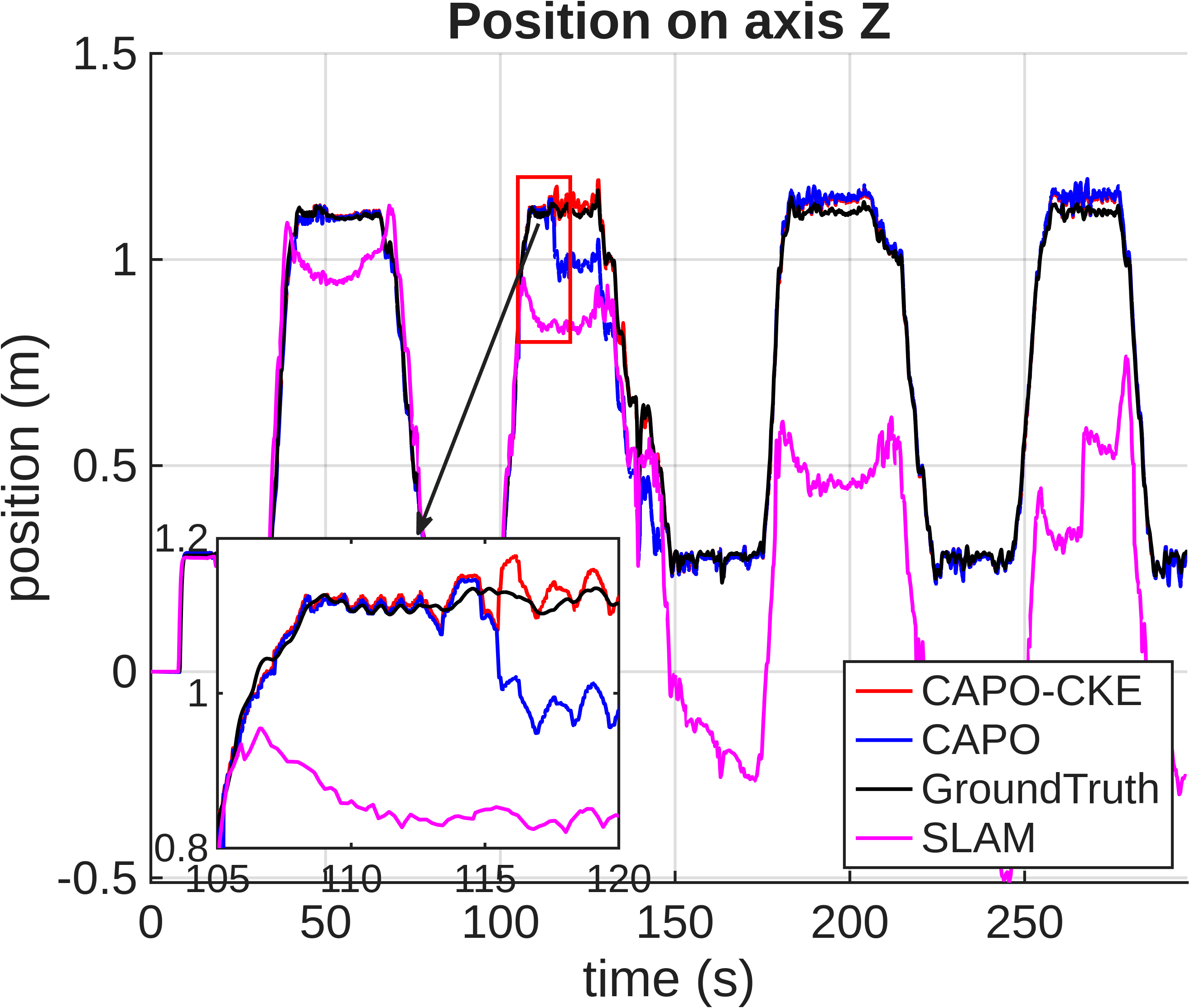}
        \caption{Estimated z-position during stair climbing.}
        \label{P4_2_8}
    \end{minipage}
    \hfill
    \begin{minipage}[b]{0.48\linewidth}
        \centering
        \includegraphics[width=1\linewidth]{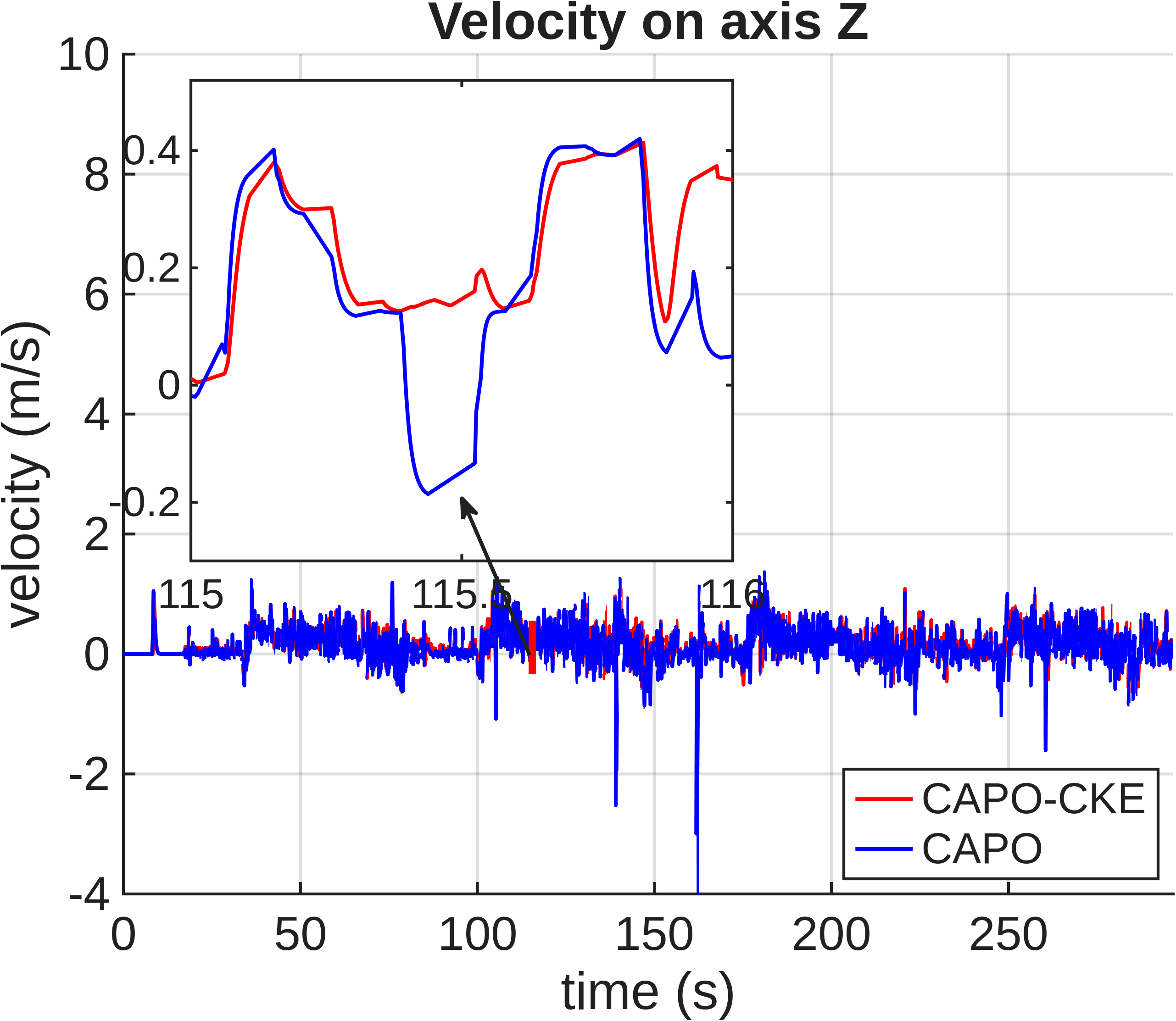}
        \caption{Estimated z-velocity during stair climbing.}
        \label{P4_2_9}
    \end{minipage}
\end{figure*}

The stair-climbing loop stresses proprioceptive velocity feedback due to repeated impacts and increased trunk vibration.
As shown in Fig.~\ref{P4_2_9}, the body velocity inferred directly from encoder-derived joint rates exhibits larger-amplitude jitter and occasional impulsive outliers.
In CAPO (without IKVel-CKF), a single pronounced spike around $t\approx 115.5$\,s corrupts the stance/landing decision and induces an erroneous foot-contact update, which subsequently biases the estimated support height and leads to a persistent degradation of the $z$ estimate.
In contrast, CAPO-CKE attenuates such outliers via IKVel-CKF (Section~IV), thereby preventing the false contact event and maintaining consistent long-horizon height tracking (Fig.~\ref{P4_2_8}--\ref{P4_2_9}).
Meanwhile, the SLAM baseline (pink curve) continues to drift in this low-texture environment, resulting in progressively worse estimates over time.

\subsection{Real-World Experiments}
\label{sec:real}

We evaluate CAPO on four quadruped platforms: a Unitree Go2 EDU and three Astrall robots,
including a point-foot platform A and two wheel-legged platforms B, C.
All experiments are conducted in a purely proprioceptive regime using IMU and motor sensing only;
no exteroceptive pose feedback is used for online drift correction.

For the Go2 planar-loop trial only, we additionally run ROS~2 \texttt{slam\_toolbox} in a mapping-only setting
with scan matching disabled, such that geometric inconsistencies in the reconstructed map directly visualize
odometry drift (rather than scan-to-map optimization effects).
For the remaining trials (including all Astrall experiments), we report the real-time estimated $(x,y,z)$ traces.

Horizontal-loop error is reported as the planar closure error $e_{xy}=\sqrt{\Delta x^2+\Delta y^2}$.
For vertical-loop tests we report the height closure error $e_z=|\Delta z|$, since long-horizon elevation consistency is the primary objective in these sequences.

A summary of the closed-loop results is provided in Table~\ref{tab:exp_summary}.

\begin{table}[htbp]
    \centering
    \caption{Closed-loop performance. Horizontal: planar closure error $e_{xy}$. Vertical: height closure error $e_z$.}
    \label{tab:exp_summary}
    \begin{tabular}{|l|c|c|}
        \hline
        \textbf{Platform} & \textbf{Horizontal loop} & \textbf{Vertical loop} \\
        \hline
        Unitree Go2 EDU & $\sim$120\,m, $2.2138$\,m & $\sim$8\,m, $|\Delta z|<0.1$\,m \\
        Astrall A (point-foot) & $\sim$200\,m, $0.1638$\,m & $\sim$15\,m, $0.219$\,m \\
        Astrall B (wheel-legged) & $\sim$200\,m, $0.2264$\,m & $\sim$15\,m, $0.199$\,m \\
        Astrall C (wheel-legged) & $\sim$700\,m, $7.68$\,m & $\sim$20\,m, $0.540$\,m \\
        \hline
    \end{tabular}
\end{table}

\subsubsection{Unitree Go2 EDU}
\label{sec:go2_real}

\begin{figure*}[t]
    \centering
    \begin{minipage}[t]{0.49\textwidth}
        \centering
        \includegraphics[width=\linewidth]{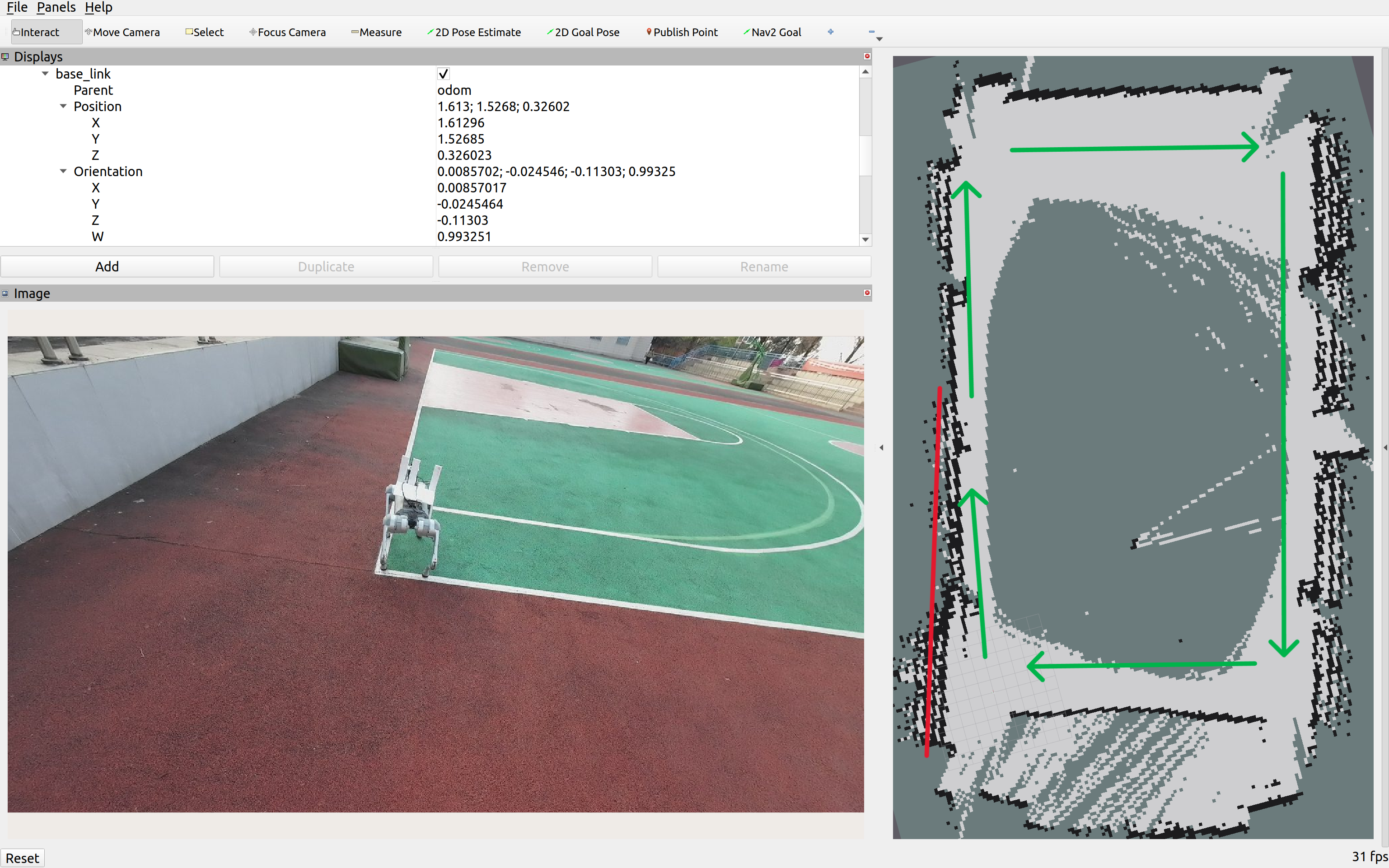}\\[-0.6ex]
        {\scriptsize (a) Mapping-only RViz view (\texttt{slam\_toolbox} with scan matching disabled).}
    \end{minipage}
    \hfill
    \begin{minipage}[t]{0.49\textwidth}
        \centering
        \includegraphics[width=\linewidth]{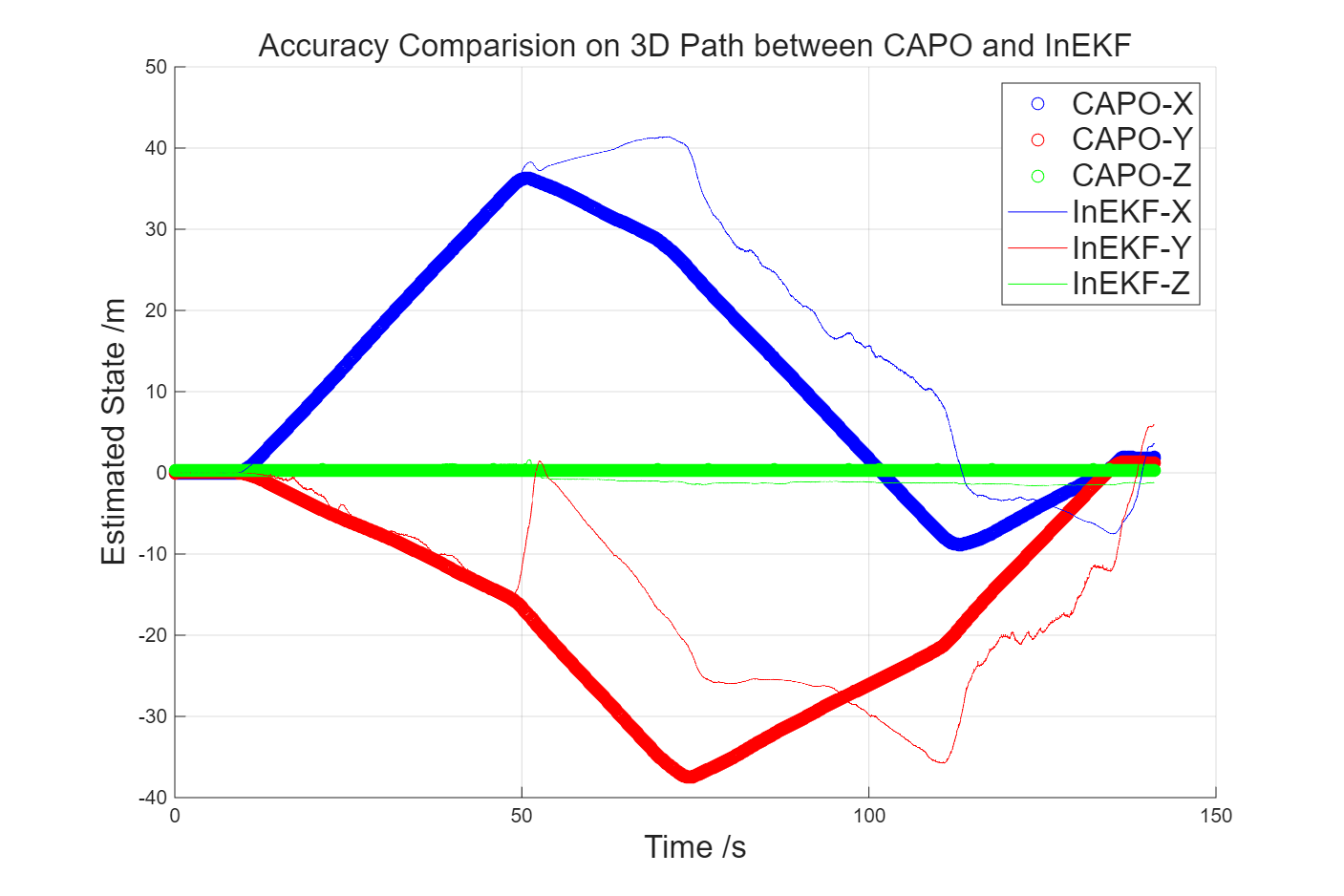}\\[-0.6ex]
        {\scriptsize (b) Real-time estimated position ($x/y/z$ in blue/red/green).}
    \end{minipage}
    \caption{Unitree Go2 EDU planar closed-loop traversal on a basketball court.}
    \label{fig:go2_flat}
\end{figure*}

\paragraph{Flat-court closed loop.}
The Go2 executes a planar closed-loop traversal on a basketball court, starting from the map corner,
following the planned trajectory, and returning to the initial area.
At loop closure, the terminal error is $\Delta x=1.61$\,m and $\Delta y=1.52$\,m,
corresponding to a planar distance error of $2.2138$\,m.
Figure~\ref{fig:go2_flat} reports both the mapping-only visualization and the real-time estimated $(x,y,z)$ trace.
The map exhibits a small but visible wall-orientation inconsistency (newly observed black segment versus the manually annotated red extension),
which is consistent with residual heading drift when scan matching is disabled.
This observation supports that yaw drift remains a major contributor to long-horizon horizontal error
in the absence of an exteroceptive heading reference.

\begin{figure*}[t]
    \centering
    \begin{minipage}[t]{0.49\textwidth}
        \centering
        \includegraphics[width=\linewidth]{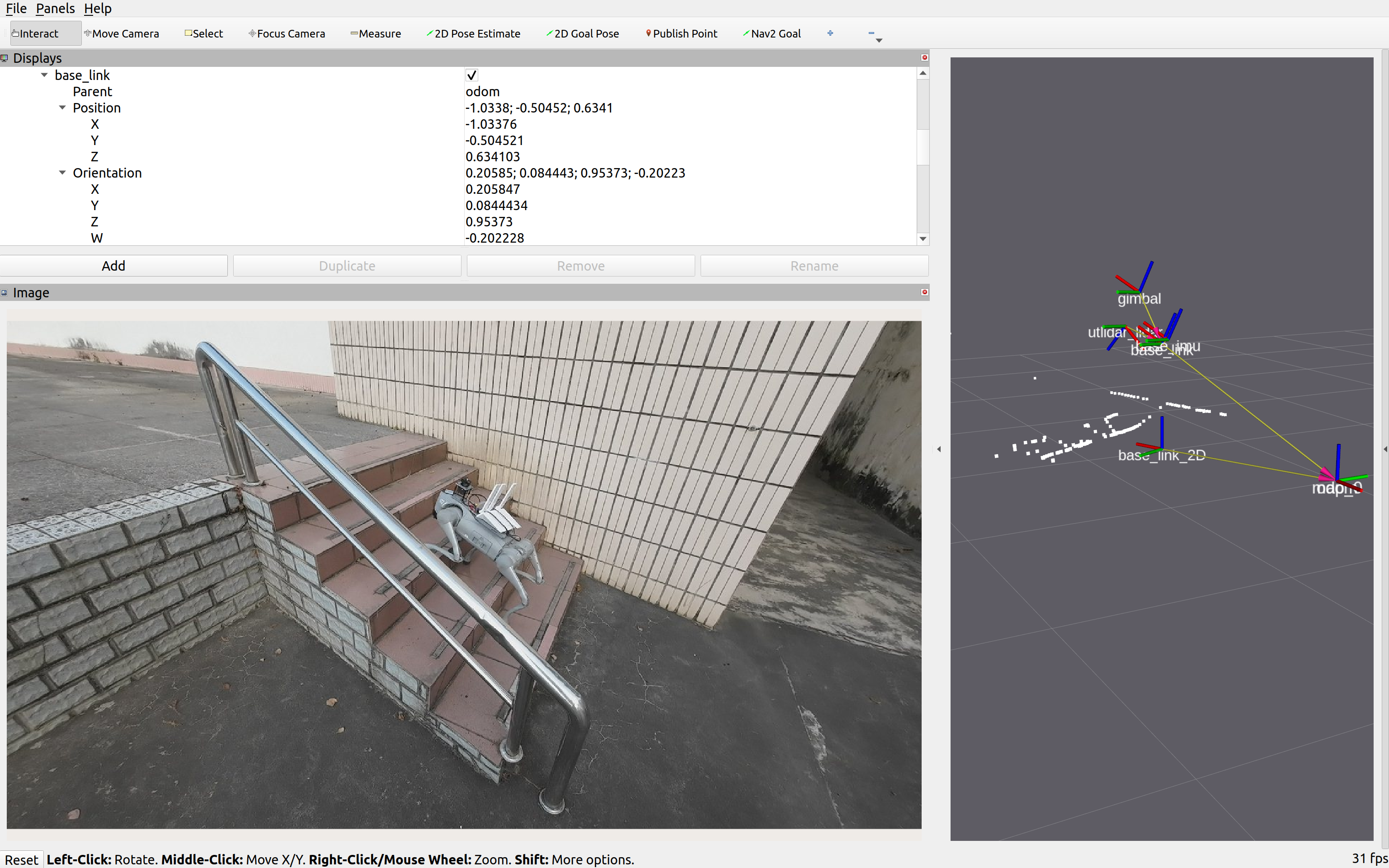}\\[-0.6ex]
        {\scriptsize (a) RViz visualization of repeated step traversal (five cycles).}
    \end{minipage}
    \hfill
    \begin{minipage}[t]{0.49\textwidth}
        \centering
        \includegraphics[width=\linewidth]{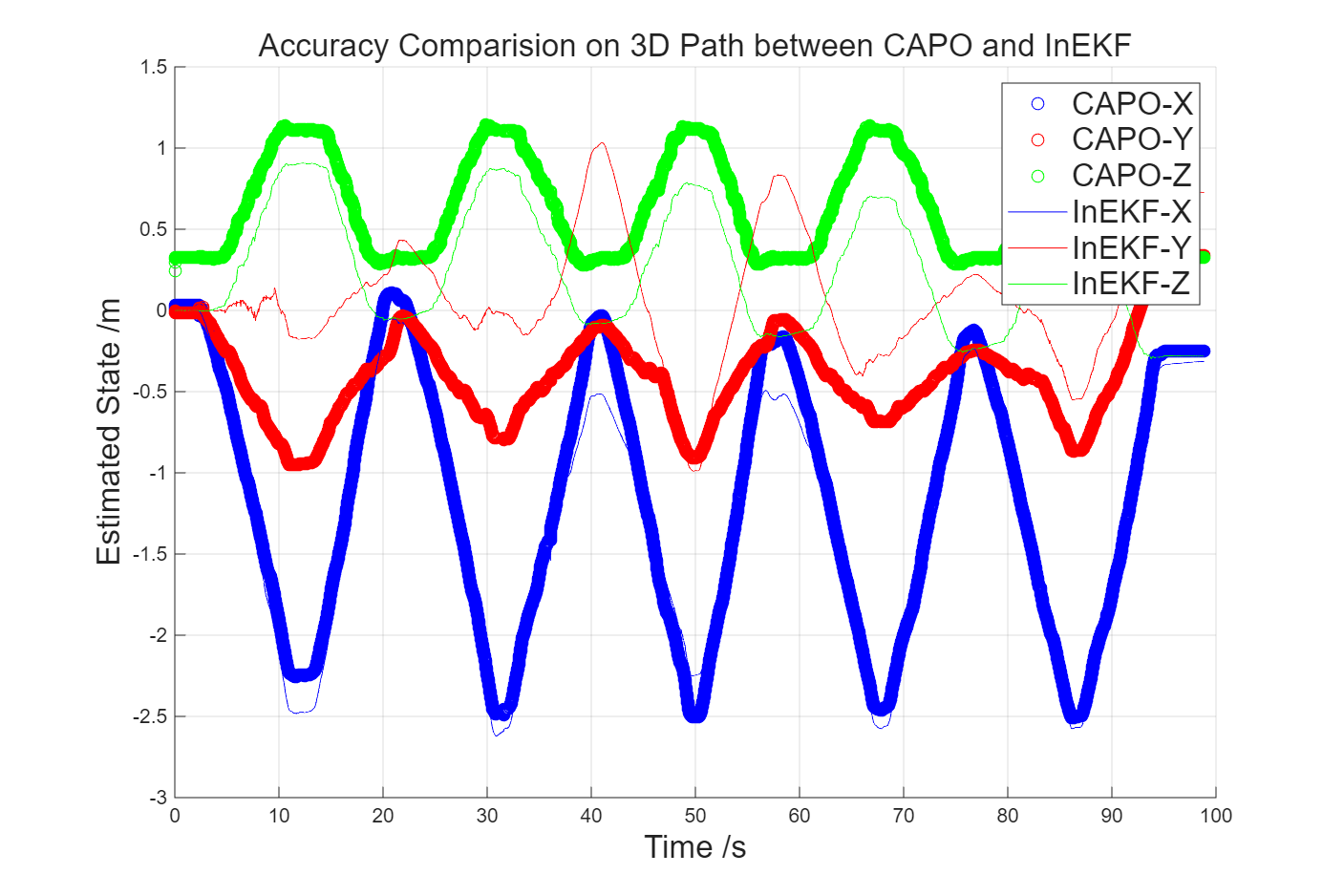}\\[-0.6ex]
        {\scriptsize (b) Real-time estimated position ($x/y/z$ in blue/red/green).}
    \end{minipage}
    \caption{Unitree Go2 EDU repeated ascent/descent over a single low step.}
    \label{fig:go2_stair}
\end{figure*}

\paragraph{Repeated step-up/step-down.}
To stress elevation consistency under repeated contact switching, the Go2 repeatedly traverses a single low step for five cycles (up/down).
Figure~\ref{fig:go2_stair} shows the RViz view and the corresponding real-time estimated $(x,y,z)$ trace.
Across the five cycles, the estimator exhibits no visually observable accumulated drift in the vertical channel at the plot scale,
i.e., $|\Delta z|<0.1$\,m upon returning to the initial region, indicating stable elevation consistency under repetitive elevation changes.

\subsubsection{Astrall Platforms (A, B, C)}
\label{sec:astrall_real}

The three Astrall platforms are equipped with higher-grade IMUs than Go2, yielding reduced inertial bias and improved heading stability,
which enables tighter closed-loop consistency in typical scenarios.
Moreover, stance detection and contact anchoring are generally more reliable on the point-foot platform A,
whereas wheel-legged platforms B and C face additional challenges from wheel slip and intermittent contact conditions.
In particular, slip is not explicitly detected or compensated in the current implementation,
and step-down motions on wheel-legged robots can exhibit brief ballistic phases that violate the ideal stationary-contact assumption,
both of which can degrade horizontal-loop closure.

\begin{figure*}[t]
    \centering
    \begin{minipage}[t]{0.49\textwidth}
        \centering
        \includegraphics[width=\linewidth]{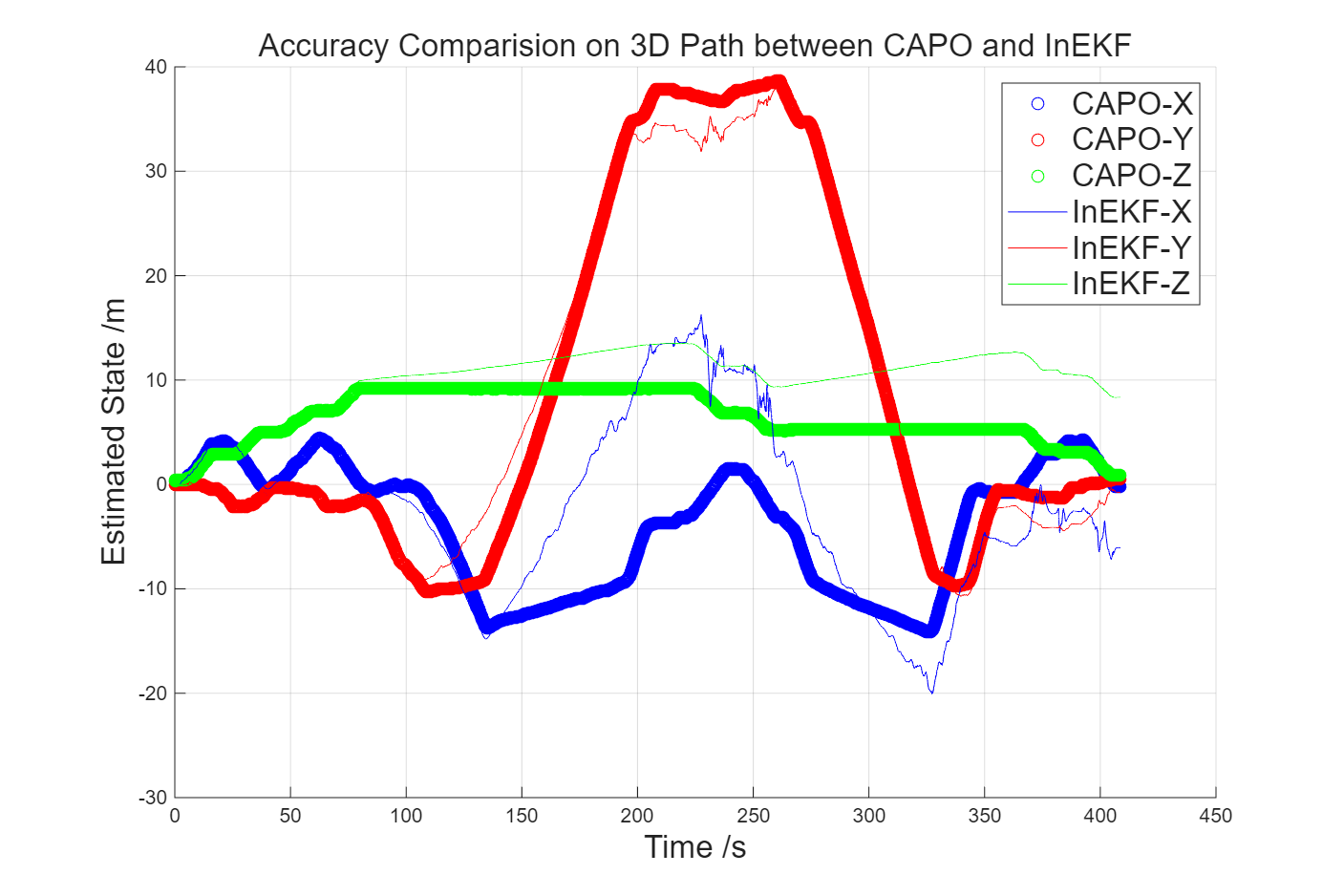}\\[-0.6ex]
        {\scriptsize (a) Astrall A (point-foot): $\sim$200\,m horizontal loop and $\sim$15\,m vertical loop.}
    \end{minipage}
    \hfill
    \begin{minipage}[t]{0.49\textwidth}
        \centering
        \includegraphics[width=\linewidth]{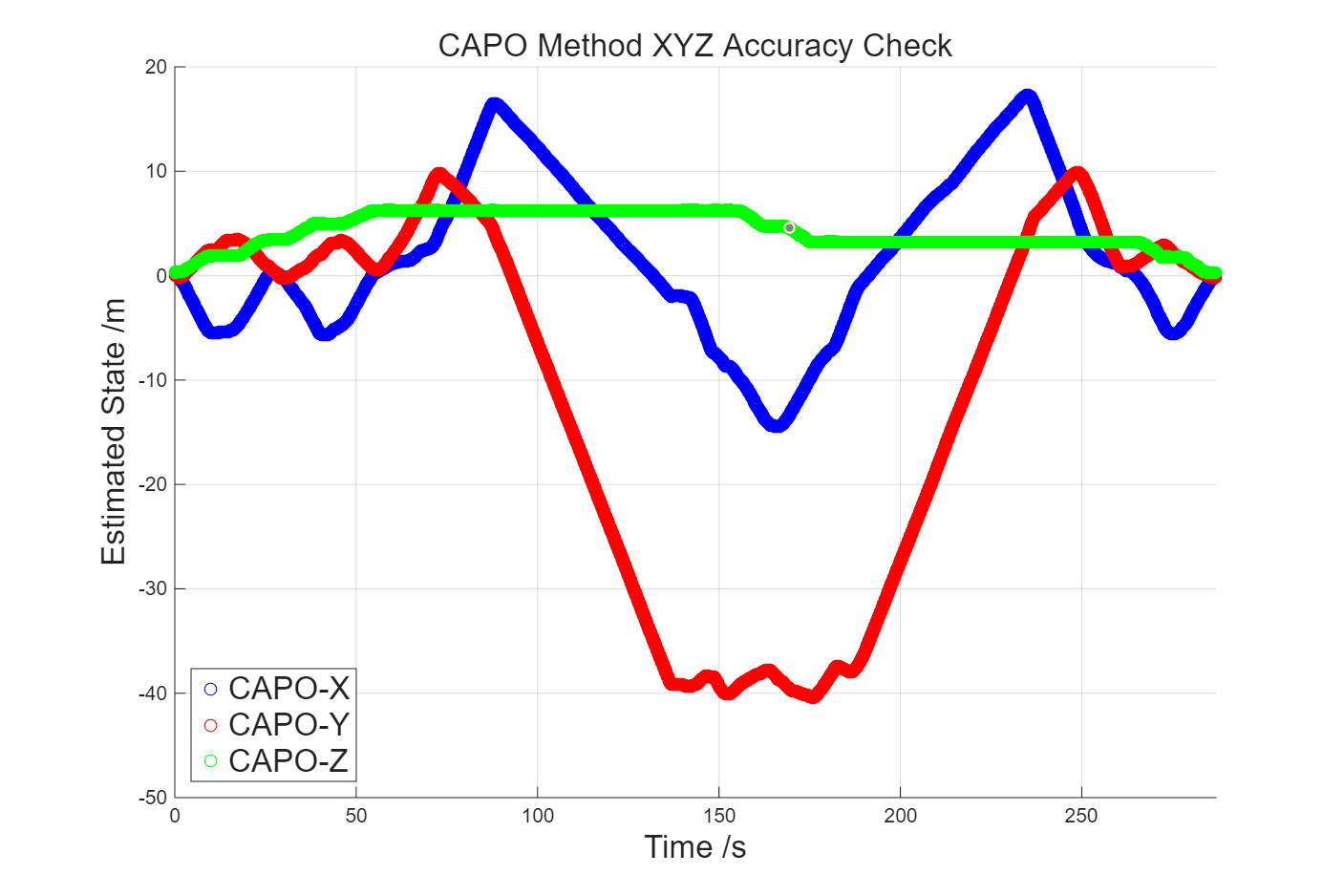}\\[-0.6ex]
        {\scriptsize (b) Astrall B (wheel-legged): $\sim$200\,m horizontal loop and $\sim$15\,m vertical loop.}
    \end{minipage}
    \caption{Astrall closed-loop trials: real-time estimated $(x,y,z)$ traces for Robot~A (MP) and Robot~B (MW).}
    \label{fig:astrall_ab}
\end{figure*}

\paragraph{Robot A vs.\ Robot B (single mixed 3D loop)}
Figure~\ref{fig:astrall_ab} reports representative real-time estimated $(x,y,z)$ traces for Robot~A and Robot~B.
Each robot executes one closed-loop trajectory that combines approximately $\sim$200\,m of horizontal travel
with an accumulated $\sim$15\,m vertical excursion, returning to the start.
Robot~A achieves $0.1638$\,m horizontal-loop error and $0.219$\,m vertical-loop error,
while Robot~B yields $0.2264$\,m and $0.199$\,m, respectively (Table~\ref{tab:exp_summary}).

\begin{figure*}[t]
    \centering
    \begin{minipage}[t]{0.49\textwidth}
        \centering
        \includegraphics[width=\linewidth]{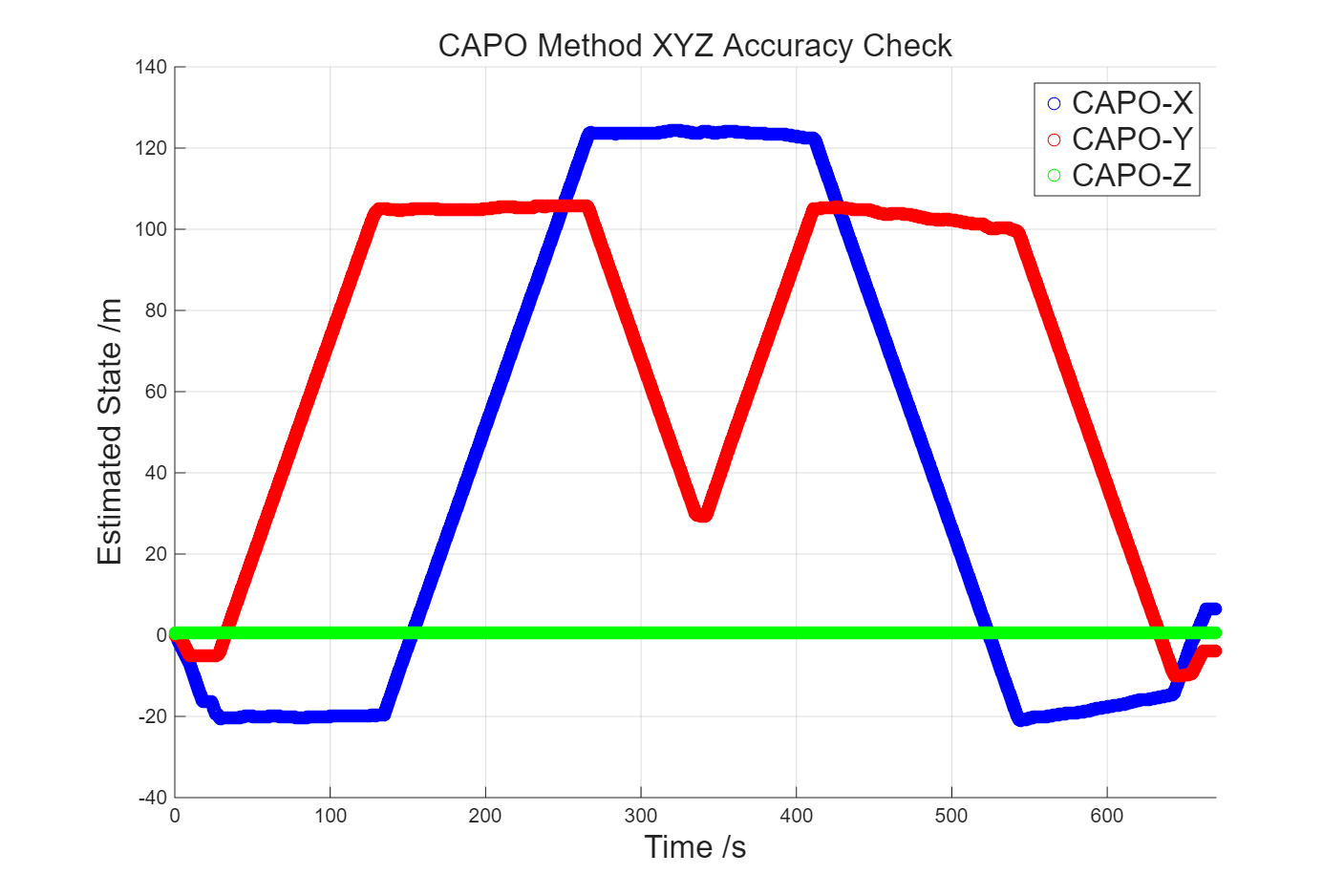}\\[-0.6ex]
        {\scriptsize (a) Astrall C: $\sim$700\,m horizontal closed loop.}
    \end{minipage}
    \hfill
    \begin{minipage}[t]{0.49\textwidth}
        \centering
        \includegraphics[width=\linewidth]{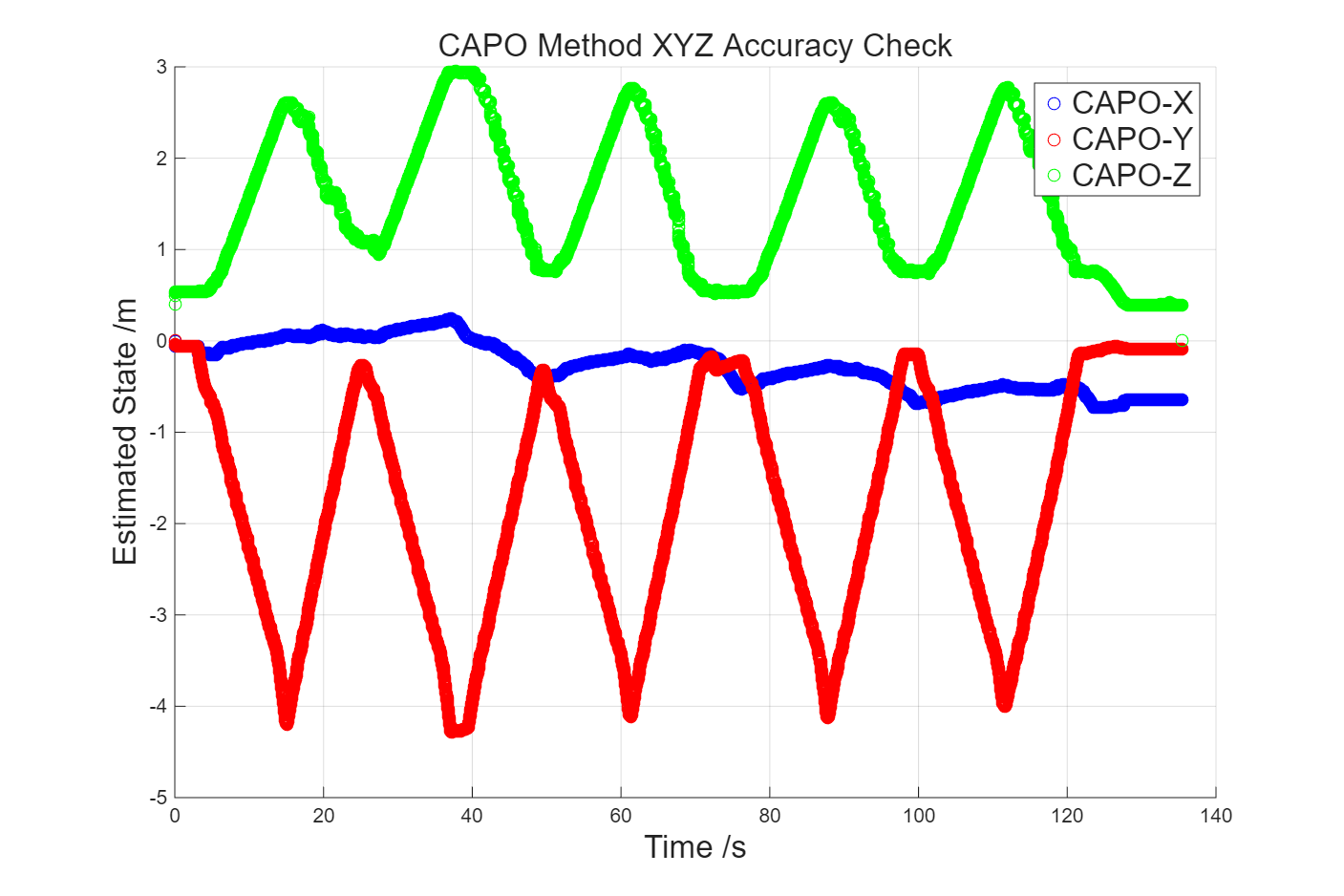}\\[-0.6ex]
        {\scriptsize (b) Astrall C: $\sim$20\,m vertical closed loop.}
    \end{minipage}
    \caption{Astrall Robot~C closed-loop trials: real-time estimated $(x,y,z)$ traces.}
    \label{fig:astrall_c}
\end{figure*}

\paragraph{Robot C (two separate trials)}
Robot~C is evaluated in two separate runs: (i) a substantially longer $\sim$700\,m horizontal closed loop,
and (ii) a $\sim$20\,m vertical loop trajectory.
Figure~\ref{fig:astrall_c} reports the corresponding real-time estimated $(x,y,z)$ traces.
The horizontal-loop error increases to $7.68$\,m, while the vertical-loop error remains $0.540$\,m (Table~\ref{tab:exp_summary}).
We attribute the larger horizontal drift primarily to slip events and contact-geometry violations that are more pronounced
for long-range wheel-legged operation, especially during step-down segments where the motion may involve brief loss of sustained rolling contact.

\paragraph{Public artifacts.}
To support reproducible evaluation, we release (i) the complete implementation and (ii) the Unitree Go2 EDU test data,
including ROS bags, CSV files, and videos.\footnote{\url{https://github.com/ShineMinxing/CAPO-LeggedRobotOdometry}}
\footnote{\url{https://github.com/ShineMinxing/CAPO-LeggedRobotOdometry/releases/tag/DataForTest}}

\subsection{Discussion and Limitations}
\label{sec:discussion}

\paragraph{Wheel slip.}
The current wheel-legged pipeline does not explicitly detect or compensate wheel--ground slip.
Slip violates the rolling/contact assumptions and is a likely source of long-horizon drift, especially over long-range runs.
Future work will incorporate slip-aware gating/weighting using kinematic--inertial inconsistency and/or wrench cues.

\paragraph{Contact detection with a fixed threshold.}
Stance selection currently relies on a constant vertical-force threshold.
This is robust for typical point-foot gaits, but can be suboptimal for wheel-legged modes with prolonged light contact, while lowering the threshold may harm stair-climbing robustness.
An adaptive or probabilistic contact model is therefore desirable.

\paragraph{Slope-aware wheel contact propagation.}
The current implementation includes a slope-aware propagation mode that fits a local support plane from reliable stance contacts and lifts the rolling direction from the horizontal plane to the estimated tangent direction.
This improves wheel-contact anchoring on moderate ramps and stairs, but its reliability depends on a sufficient contact set, stable contact timing, and limited slip.
Future work will therefore combine the existing local support-plane propagation with explicit slip detection and uncertainty-adaptive contact weighting.

\paragraph{Supplementary clarification on biped applicability.}
CAPO is contact-set driven and thus extends to bipeds by operating on a time-varying set of contacting end-effectors.
A demonstration video shows a quadruped switching between two-foot and four-foot morphologies while completing a closed loop without scan matching.\footnote{\url{https://www.bilibili.com/video/BV1UtQfYJExu}}

\paragraph{Supplementary yaw-drift compensation and IMU-free degradation.}
We provide a video showing that the kinematics-based yaw correction arrests IMU yaw drift during prolonged standing and remains stable under short in-place rotations.\footnote{\url{https://www.bilibili.com/video/BV1frykB8ETV}}
When IMU yaw constraints are disabled, the kinematics-only heading reference degrades gracefully, with empirical errors of $\sim 10^\circ$ per closed turn for short-step turning and $\sim 30^\circ$ per closed turn for long-step turning, suggesting residual sensitivity to unmodeled attitude coupling (notably roll/pitch) and contact compliance.

\section{Conclusion}
\label{sec:conclusion}

This paper presented \emph{CAPO}, a purely proprioceptive, contact-anchored odometry framework for legged and wheel-legged robots using only IMU and motor measurements.
The estimator treats stance end-effectors as intermittent world-frame anchors by recording footfall positions at touchdown and enforcing contact-anchored kinematic consistency during stance.
To stabilize long-horizon elevation, we introduced a lightweight support-plane height correction that clusters touchdown heights with time-decayed confidence.
For wheel-legged platforms, we incorporated shank-motion-compensated rolling-contact propagation, including lateral roll compensation and a slope-aware rolling direction, to propagate wheel contacts consistently during stance.
To mitigate encoder-induced impulsive artifacts in hip--foot velocity, we further integrated an optional IKVel-CKF, yielding the \emph{CAPO-CKE} variant with substantially smoother velocity feedback for fusion.
Finally, we proposed a kinematics-based yaw correction based on multi-contact geometric consistency, which arrests IMU yaw drift during prolonged standing and provides a graceful kinematics-only fallback when IMU yaw constraints are degraded.

We evaluated the approach in both physics simulation and real-robot trials.
In simulation, CAPO and CAPO-CKE significantly outperformed the SLAM baseline in low-texture settings, particularly in the vertical channel, and CAPO-CKE improved robustness against occasional encoder-driven velocity outliers during stair climbing.
On real hardware, we validated the estimator on four quadruped platforms (Unitree Go2 EDU and three Astrall robots) using closed-loop trajectories with horizontal and vertical motion.
The results demonstrate that contact anchoring with touchdown footfall records can provide accurate long-horizon proprioceptive odometry in practice, while residual horizontal drift remains most sensitive to heading errors and contact-model violations (e.g., slip).

There remain several limitations and opportunities for improvement.
Future work will focus on (i) explicit slip detection and down-weighting for wheel-legged contacts, (ii) adaptive contact classification beyond fixed-threshold wrench gating to handle prolonged light contacts without sacrificing stair robustness, and (iii) improving the existing slope-aware wheel-contact propagation with more reliable local support-plane estimation and uncertainty-aware terrain association.
In addition, reducing the runtime overhead of IKVel-CKF (e.g., by conditional enabling, lower-rate updates, or more efficient sigma-point implementations) would improve deployability on resource-constrained onboard computers.

\bibliographystyle{IEEEtran}
\bibliography{NSEQRBFRbib}

\appendices
\section{Leg Kinematics and Wrench Estimation}
\label{app:leg_kin_wrench}

This appendix summarizes the kinematic and wrench models used to compute end-effector position, velocity, and contact force from joint measurements.
The released implementation uses a configurable transform chain for each contact limb; the closed-form 3-DoF model below is retained as an analytically transparent specialization for the platforms evaluated in this paper.

\subsection{Transform-Chain Implementation}
Let node $n$ in leg $i$ have parent $\pi(n)$, fixed translation $\mathbf{t}_{i,n}$, fixed rotation $\mathbf{R}^{0}_{i,n}$, joint axis $\mathbf{a}_{i,n}$, and joint angle $q_{i,n}$.
The world-frame node transform is propagated recursively as
\begin{equation}
\label{eq:tf_chain_pos}
\mathbf{o}^{W}_{i,n}=\mathbf{o}^{W}_{i,\pi(n)}+\mathbf{R}^{W}_{i,\pi(n)}\mathbf{t}_{i,n},
\end{equation}
\begin{equation}
\label{eq:tf_chain_rot}
\mathbf{R}^{W}_{i,n}=\mathbf{R}^{W}_{i,\pi(n)}\mathbf{R}^{0}_{i,n}\exp\big(q_{i,n}[\mathbf{a}_{i,n}]_\times\big),
\end{equation}
where the body pose provides the root transform when $\pi(n)=-1$.
The end-effector position is obtained by applying the terminal fixed offset to the final node.
For every active joint, the world-frame geometric Jacobian column is
\begin{equation}
\label{eq:chain_jacobian}
\mathbf{J}^{W}_{i,n}=\mathbf{a}^{W}_{i,n}\times\left(\mathbf{p}^{W}_{\mathrm{ee},i}-\mathbf{o}^{W}_{i,n}\right),
\end{equation}
which gives the default leg velocity used in \eqref{eq:vel_obs_leg}.
The same Jacobian is used for torque-based wrench estimation through \eqref{eq:force_impl_appendix}.

\subsection{Parameters and Modeling Convention}
The released implementation uses the transform-chain model above.
For analytical interpretation and reproducibility of the quadruped point-foot specialization, we also give a 3-DoF closed-form leg model with joint angles $\mathbf{q}=[q_1,q_2,q_3]^T$ and joint velocities $\dot{\mathbf{q}}=[\dot q_1,\dot q_2,\dot q_3]^T$.
We use a side sign $s\in\{+1,-1\}$ to distinguish left/right legs.

Link parameters are:
\begin{itemize}[leftmargin=*]
\item $L_{\mathrm{hip}}$: hip lateral offset,
\item $L_{\mathrm{thigh}}$: thigh link length,
\item $L_{\mathrm{calf}}$: calf/shank effective length,
\item $r_w$: wheel radius (wheel-legged only).
\end{itemize}

\textbf{Point-foot convention (matching our code).} For point-foot robots, the foot-end radius is treated as an effective extension of the shank, hence it is absorbed into $L_{\mathrm{calf}}$ and we set $r_w=0$.
\textbf{Wheel-legged convention.} For wheel-legged robots, $L_{\mathrm{calf}}$ denotes the link length to the wheel axis, and $r_w>0$ is the wheel radius.

Define the trigonometric shorthand
\begin{equation}
c_i=\cos q_i,\quad s_i=\sin q_i,\quad
c_{23}=\cos(q_2+q_3)=c_2c_3-s_2s_3,\quad
s_{23}=\sin(q_2+q_3)=s_2c_3+c_2s_3.
\end{equation}

\subsection{Hip-to-End-Effector Relative Position (Body Frame)}
\label{app:fk_pos}
The hip-to-end-effector relative position (expressed in the body frame) is
\begin{equation}
\label{eq:fk_pos_impl}
\mathbf{r}^{B}(\mathbf{q})=
\begin{bmatrix}
x\\y\\z
\end{bmatrix}
=
\begin{bmatrix}
-\big(L_{\mathrm{calf}}\,s_{23}+L_{\mathrm{thigh}}\,s_2\big)\\
s\,L_{\mathrm{hip}}\,c_1 + (L_{\mathrm{calf}}+r_w)\,s_1\,c_{23} + L_{\mathrm{thigh}}\,c_2\,s_1\\
s\,L_{\mathrm{hip}}\,s_1 - L_{\mathrm{calf}}\,c_1\,c_{23} - L_{\mathrm{thigh}}\,c_1\,c_2 + r_w
\end{bmatrix}.
\end{equation}
The body-frame end-effector position is obtained by adding the hip mounting offset
\begin{equation}
\mathbf{p}^{B}_{\mathrm{ee}}=\mathbf{p}^{B}_{\mathrm{hip}}+\mathbf{r}^{B}(\mathbf{q}).
\end{equation}

\subsection{Hip-to-End-Effector Relative Velocity (Body Frame)}
\label{app:fk_vel}
Differentiating \eqref{eq:fk_pos_impl} yields the relative linear velocity
\begin{equation}
\label{eq:fk_vel_impl}
\mathbf{v}^{B}(\mathbf{q},\dot{\mathbf{q}})=
\begin{bmatrix}
v_x\\v_y\\v_z
\end{bmatrix}
=
\begin{bmatrix}
-\big((L_{\mathrm{calf}}c_{23}+L_{\mathrm{thigh}}c_2)\dot q_2 + (L_{\mathrm{calf}}c_{23})\dot q_3\big)\\
\big((L_{\mathrm{calf}}+r_w)c_1c_{23}+L_{\mathrm{thigh}}c_1c_2-sL_{\mathrm{hip}}s_1\big)\dot q_1 \\
\quad +\big(-(L_{\mathrm{calf}}+r_w)s_1s_{23}-L_{\mathrm{thigh}}s_1s_2\big)\dot q_2
+\big(-(L_{\mathrm{calf}}+r_w)s_1s_{23}\big)\dot q_3\\
\big(L_{\mathrm{calf}}s_1c_{23}+L_{\mathrm{thigh}}c_2s_1+sL_{\mathrm{hip}}c_1\big)\dot q_1 \\
\quad +\big(L_{\mathrm{calf}}c_1s_{23}+L_{\mathrm{thigh}}c_1s_2\big)\dot q_2
+\big(L_{\mathrm{calf}}c_1s_{23}\big)\dot q_3
\end{bmatrix}.
\end{equation}

\subsection{Geometric Jacobian}
\label{app:jacobian}
The geometric Jacobian $\mathbf{J}(\mathbf{q})\in\mathbb{R}^{3\times 3}$ is defined by
\begin{equation}
\mathbf{v}^{B}(\mathbf{q},\dot{\mathbf{q}})=\mathbf{J}(\mathbf{q})\,\dot{\mathbf{q}}.
\end{equation}
From \eqref{eq:fk_pos_impl}, the Jacobian entries are
\begin{equation}
\label{eq:J_entries}
\mathbf{J}(\mathbf{q})=
\begin{bmatrix}
0 &
-(L_{\mathrm{calf}}c_{23}+L_{\mathrm{thigh}}c_2) &
-(L_{\mathrm{calf}}c_{23})\\
(L_{\mathrm{calf}}+r_w)c_1c_{23}+L_{\mathrm{thigh}}c_1c_2-sL_{\mathrm{hip}}s_1 &
-(L_{\mathrm{calf}}+r_w)s_1s_{23}-L_{\mathrm{thigh}}s_1s_2 &
-(L_{\mathrm{calf}}+r_w)s_1s_{23}\\
L_{\mathrm{calf}}s_1c_{23}+L_{\mathrm{thigh}}c_2s_1+sL_{\mathrm{hip}}c_1 &
L_{\mathrm{calf}}c_1s_{23}+L_{\mathrm{thigh}}c_1s_2 &
L_{\mathrm{calf}}c_1s_{23}
\end{bmatrix}.
\end{equation}

\subsection{End-Effector Force Estimation from Joint Torques}
\label{app:wrench}
Let $\boldsymbol{\tau}=[\tau_1,\tau_2,\tau_3]^T$ denote the measured joint torques.
Under the quasi-static mapping
\begin{equation}
\boldsymbol{\tau}=\mathbf{J}^T(\mathbf{q})\,\mathbf{f}^{B},
\end{equation}
we compute the end-effector force in the body frame by
\begin{equation}
\label{eq:force_impl_appendix}
\mathbf{f}^{B}=(\mathbf{J}\mathbf{J}^T)^{-1}\mathbf{J}\boldsymbol{\tau},
\end{equation}
which is equivalent to $\mathbf{f}^{B}=\mathbf{J}^{-T}\boldsymbol{\tau}$ when $\mathbf{J}$ is invertible.
In the transform-chain implementation, the Jacobian can be constructed directly in the world frame, yielding the same virtual-work relation without a separate body-frame rotation.
The vertical component of the resulting force is used for stance gating in \eqref{eq:contact_gate2}.

\section{IKVel Inverse-Kinematics Measurement Model and Cubature Kalman Filter}
\label{app:ikvel_ckf}

This appendix summarizes the inverse-kinematics measurement model and the CKF recursion used by IKVel-CKF (Estimator1003 in our implementation).

\subsection{State, Process Model, and Side Constraint}
For end-effector $i$, define the hip-to-end-effector Cartesian state
\begin{equation}
\mathbf{x}_{i,k}=
\begin{bmatrix}
x_{i,k} & y_{i,k} & z_{i,k} & v_{x,i,k} & v_{y,i,k} & v_{z,i,k}
\end{bmatrix}^{\!\top}.
\end{equation}
A constant-velocity prior is used:
\begin{equation}
\label{eq:ikvel_process}
\mathbf{x}_{i,k+1} =
\underbrace{\begin{bmatrix}
\mathbf{I}_3 & \Delta t_k \mathbf{I}_3\\
\mathbf{0}_3 & \mathbf{I}_3
\end{bmatrix}}_{\mathbf{F}(\Delta t_k)}
\mathbf{x}_{i,k} + \mathbf{w}_{i,k},
\end{equation}
where $\Delta t_k=t_{k+1}-t_k$ is truncated for robustness to timestamp anomalies (in the implementation, $\Delta t_k$ is set to $0$ when $|\Delta t_k|$ exceeds a small bound), and $\mathbf{w}_{i,k}\sim\mathcal{N}(\mathbf{0},\mathbf{Q})$.
To enforce a consistent kinematic branch for left/right legs, we apply a side constraint on the lateral coordinate,
\begin{equation}
\label{eq:ikvel_side}
y_{i,k} \leftarrow s_i |y_{i,k}|,\qquad s_i\in\{+1,-1\},
\end{equation}
where $s_i$ denotes the leg side sign.

\subsection{Inverse-Kinematics Measurement Model}
The measurement is the joint configuration and joint velocity:
\begin{equation}
\label{eq:ikvel_meas_app}
\mathbf{z}_{i,k} \triangleq
\begin{bmatrix}
\mathbf{q}_{i,k}\\
\dot{\mathbf{q}}_{i,k}
\end{bmatrix}
=
\mathbf{h}(\mathbf{x}_{i,k}) + \mathbf{v}_{i,k},\qquad
\mathbf{v}_{i,k}\sim\mathcal{N}(\mathbf{0},\mathbf{R}).
\end{equation}
Let $L_{\mathrm{hip}}$, $L_{\mathrm{thigh}}$, $L_{\mathrm{calf}}$ be link lengths and let $r_{\mathrm{ee}}$ be the modeled end-effector radius (absorbed into the effective shank length for point feet, explicit for wheels).
Define $L_2 \triangleq L_{\mathrm{calf}} + r_{\mathrm{ee}}$.
Given $(x,y,z)$, the analytic inverse kinematics computes $(\theta_1,\theta_2,\theta_3)$ as follows.
First solve the ab/adduction angle $\theta_1$ in the $(y,z)$ plane:
\begin{equation}
\label{eq:ik_theta1}
\theta_1
= s_i\arcsin\!\left(
\frac{2L_{\mathrm{hip}}z + \sqrt{\epsilon + 4L_{\mathrm{hip}}^2 z^2 -4(z^2+y^2)\big(L_{\mathrm{hip}}^2-y^2\big)}}
{2(z^2+y^2)}
\right),
\end{equation}
where $\epsilon>0$ is a small constant added for numerical robustness.
Then remove the hip offset:
\begin{equation}
\label{eq:ik_yzbar}
\bar z = z - s_i L_{\mathrm{hip}}\sin\theta_1,\qquad
\bar y = y - s_i L_{\mathrm{hip}}\cos\theta_1.
\end{equation}
Let $\bar r=\sqrt{\bar y^2+\bar z^2}$ and $r=\sqrt{\bar r^2+x^2}$.
The remaining planar angles are
\begin{align}
\label{eq:ik_theta23}
\theta_3 &= -\pi + \arccos\!\left(\frac{L_{\mathrm{thigh}}^2 + L_2^2 - r^2}{2L_{\mathrm{thigh}}L_2}\right),\\
\theta_2 &= \arctan2(x,\bar r) + \arccos\!\left(\frac{r^2 + L_{\mathrm{thigh}}^2 - L_2^2}{2rL_{\mathrm{thigh}}}\right).
\end{align}

To predict joint velocities from $(v_x,v_y,v_z)$, we use the differential relationship
\begin{equation}
\label{eq:ik_diff}
\mathbf{v} = \mathbf{J}(\boldsymbol{\theta})\,\dot{\boldsymbol{\theta}},
\qquad
\dot{\boldsymbol{\theta}} = \mathbf{J}(\boldsymbol{\theta})^{-1}\mathbf{v},
\end{equation}
where $\boldsymbol{\theta}=[\theta_1,\theta_2,\theta_3]^\top$ and $\mathbf{v}=[v_x,v_y,v_z]^\top$.
With $c_j=\cos\theta_j$, $s_j=\sin\theta_j$, $c_{23}=\cos(\theta_2+\theta_3)$, $s_{23}=\sin(\theta_2+\theta_3)$, the Jacobian used in the implementation is
\begin{equation}
\label{eq:ik_jacobian}
\mathbf{J}(\boldsymbol{\theta}) =
\begin{bmatrix}
0 & L_2 c_{23} + L_{\mathrm{thigh}} c_2 & L_2 c_{23} \\
- s_i L_{\mathrm{hip}} s_1 + L_2 c_1 c_{23} + L_{\mathrm{thigh}} c_2 c_1
& -L_2 s_1 s_{23} - L_{\mathrm{thigh}} s_1 s_2
& -L_2 s_1 s_{23} \\
 s_i L_{\mathrm{hip}} c_1 + L_2 s_1 c_{23} + L_{\mathrm{thigh}} c_2 s_1
& \ \ L_2 c_1 s_{23} + L_{\mathrm{thigh}} c_1 s_2
& \ \ L_2 c_1 s_{23}
\end{bmatrix}.
\end{equation}
The observation function is therefore
\begin{equation}
\label{eq:ik_h}
\mathbf{h}(\mathbf{x}) =
\begin{bmatrix}
\theta_1 & \theta_2 & \theta_3 &
\dot\theta_1 & \dot\theta_2 & \dot\theta_3
\end{bmatrix}^{\!\top},
\qquad
\dot{\boldsymbol{\theta}} = \mathbf{J}(\boldsymbol{\theta})^{-1}\mathbf{v}.
\end{equation}

\subsection{Cubature Kalman Filter Recursion}
Let $n=\mathrm{dim}(\mathbf{x})$ and let $(\hat{\mathbf{x}}_k,\mathbf{P}_k)$ be the prior estimate.
CKF constructs $2n$ cubature points using the Cholesky factor $\mathbf{S}_k$ such that $\mathbf{P}_k=\mathbf{S}_k\mathbf{S}_k^\top$:
\begin{equation}
\label{eq:ckf_points}
\boldsymbol{\chi}^{(j)}_k = \hat{\mathbf{x}}_k + \sqrt{n}\,\mathbf{S}_k \mathbf{e}_j,\qquad
\boldsymbol{\chi}^{(j+n)}_k = \hat{\mathbf{x}}_k - \sqrt{n}\,\mathbf{S}_k \mathbf{e}_j,
\end{equation}
with equal weights $w^{(m)}=\frac{1}{2n}$.
The points are propagated through the process model \eqref{eq:ikvel_process} to obtain predicted mean and covariance:
\begin{align}
\bar{\mathbf{x}}_{k}^{-} &= \sum_{m=1}^{2n} w^{(m)}\,\mathbf{f}\!\left(\boldsymbol{\chi}^{(m)}_k\right),\\
\mathbf{P}_{k}^{-} &= \mathbf{Q} + \sum_{m=1}^{2n} w^{(m)}\big(\mathbf{f}(\boldsymbol{\chi}^{(m)}_k)-\bar{\mathbf{x}}_{k}^{-}\big)\big(\mathbf{f}(\boldsymbol{\chi}^{(m)}_k)-\bar{\mathbf{x}}_{k}^{-}\big)^\top.
\end{align}
Similarly, the measurement mean and covariances are computed via \eqref{eq:ik_h}:
\begin{align}
\bar{\mathbf{z}}_{k} &= \sum_{m=1}^{2n} w^{(m)}\,\mathbf{h}\!\left(\boldsymbol{\chi}^{(m)-}_k\right),\\
\mathbf{P}_{zz} &= \mathbf{R} + \sum_{m=1}^{2n} w^{(m)}\big(\mathbf{h}(\boldsymbol{\chi}^{(m)-}_k)-\bar{\mathbf{z}}_{k}\big)\big(\mathbf{h}(\boldsymbol{\chi}^{(m)-}_k)-\bar{\mathbf{z}}_{k}\big)^\top,\\
\mathbf{P}_{xz} &= \sum_{m=1}^{2n} w^{(m)}\big(\boldsymbol{\chi}^{(m)-}_k-\bar{\mathbf{x}}_{k}^{-}\big)\big(\mathbf{h}(\boldsymbol{\chi}^{(m)-}_k)-\bar{\mathbf{z}}_{k}\big)^\top.
\end{align}
The gain and posterior update are
\begin{equation}
\mathbf{K}_k = \mathbf{P}_{xz}\mathbf{P}_{zz}^{-1},\qquad
\hat{\mathbf{x}}_{k}^{+} = \bar{\mathbf{x}}_{k}^{-} + \mathbf{K}_k(\mathbf{z}_k-\bar{\mathbf{z}}_{k}),\qquad
\mathbf{P}_{k}^{+} = \mathbf{P}_{k}^{-} - \mathbf{K}_k\mathbf{P}_{zz}\mathbf{K}_k^\top.
\end{equation}
This is the CKF recursion implemented by Estimator1003, using equal-weight spherical--radial cubature points and Cholesky factorization for numerical stability.

\end{document}